\documentclass[authoryear,preprint,12pt]{elsarticle}
\usepackage{amsmath,amssymb,amsthm,amsfonts,fourier,booktabs,blkarray,xparse,soul,comment,url,nicefrac,xcolor,bm,tikz,pgfplots,subfigure} 
\usepackage[ruled,linesnumbered]{algorithm2e}
\usepackage{hyperref,url}
  \hypersetup{  
    bookmarksnumbered
  }
  \hypersetup{  
    breaklinks=false,
    pdfborderstyle={/S/U/W 0},
    citebordercolor=.235 .702 .443,
    urlbordercolor=.255 .412 .882,
    linkbordercolor=.804 .149 .149,
  }
  \hypersetup{ 
    pdfauthor={},
    pdftitle={},
    pdfkeywords={}
}
\definecolor{myblueR2}{RGB}{0, 0, 0}
\definecolor{myblue}{RGB}{0, 0, 0}
\newtheorem{example}{Example}
\newtheorem{theorem}{Theorem}
\newtheorem{lemma}{Lemma}

\newtheorem{corollary}{Corollary}

\pgfplotsset{compat=1.17} 
\usetikzlibrary{arrows,plotmarks,decorations.markings,trees,shapes,pgfplots.groupplots,automata,circuits}
\usetikzlibrary{arrows,plotmarks,decorations.markings,trees,shapes}
\usepgfplotslibrary{statistics}
\tikzset{dot/.style = {circle, fill, minimum size=#1,inner sep=0pt, outer sep=0pt, fill, circle},dot/.default = 6pt}
\tikzset{dot2/.style = {circle, fill, color=black!40,minimum size=6pt,inner sep=0pt, outer sep=0pt, fill, circle}}
\tikzstyle{a}=[->,>=stealth,dashed]
\tikzstyle{a2}=[->,>=stealth]
\tikzstyle{a3}=[<->,>=stealth]
\tikzstyle{nodo}=[ellipse,draw=black!100,fill=black!0,line width=.7pt,minimum width=1.0cm,minimum height=0.8cm,text width=1cm,text centered]
\tikzstyle{nodo2}=[ellipse,draw=black!100,fill=black!10,line width=.7pt,minimum width=1.0cm,minimum height=0.8cm,text width=1cm,text centered]
\tikzstyle{nodo3}=[ellipse,draw=black!100,fill=black!30,line width=.7pt,minimum width=1.0cm,minimum height=0.8cm,text width=1cm,text centered]
\tikzstyle{Qnodo}=[ellipse,draw=black!100,fill=black!10,line width=.7pt,minimum width=1.2cm,minimum height=.7cm]
\tikzstyle{arco}=[draw=black!80,line width=.7pt, postaction={decorate}, decoration={markings,mark=at position 1.0 with {\arrow[ draw=black!80,line width=.7pt]{>}}}]
\tikzstyle{decision} = [rectangle, draw, fill=black!100,text=white, text width=4.5em, text badly centered, node distance=3cm, minimum height=3em]
\tikzstyle{block} = [rectangle, draw, fill=blue!20, text width=5em, text centered, rounded corners, minimum height=3em]
\tikzstyle{line} = [draw, -latex']
\tikzstyle{cloud} = [draw, ellipse,fill=red!20, node distance=3cm, minimum height=2em]

\pgfplotsset{legend image with text/.style={
legend image code/.code={%
\node[anchor=center] at (0.3cm,0cm) {#1};}},}

\newcommand{\twoFone}{\ensuremath{{}_2F_1}}
\journal{International Journal of Approximate Reasoning}
\begin{document}
\begin{frontmatter}
\title{Efficient Computation of Counterfactual Bounds}
\author[IDSIA]{Marco Zaffalon}
\ead{zaffalon@idsia.ch}
\author[IDSIA]{Alessandro Antonucci\corref{cor1}}
\ead{alessandro.antonucci@idsia.ch}
\cortext[cor1]{Corresponding author}
\author[UAL]{Rafael Cabañas}
\ead{rcabanas@ual.es}
\author[IDSIA]{David Huber}
\ead{david.huber@idsia.ch}
\author[IDSIA]{Dario Azzimonti}
\ead{dario.azzimonti@idsia.ch}
\address[IDSIA]{IDSIA, Lugano (Switzerland)}
\address[UAL]{Department of Mathematics, University of Almería, Almería (Spain)}
\tnoteref{label1}
\begin{abstract}
We assume to be given structural equations over discrete variables inducing a directed acyclic graph, namely, a \emph{structural causal model}, together with data about its internal nodes. The question we want to answer is how can we compute bounds for partially identifiable counterfactual queries from such an input. We start by giving a map from structural casual models to \emph{credal networks}. This allows us to
compute exact counterfactual bounds via algorithms for credal nets on a subclass of structural causal models. Exact computation is going to be inefficient in general given that, as we show, causal inference is NP-hard even on polytrees. We target then approximate bounds via a causal EM scheme. We evaluate their accuracy by providing credible intervals on the quality of the approximation; we show through a synthetic benchmark that the EM scheme delivers accurate results in a fair number of runs. In the course of the discussion, we also point out what seems to be a neglected limitation to the trending idea that counterfactual bounds can be computed without knowledge of the structural equations.
We also present a real case study on palliative care to show how our algorithms can readily be used for practical purposes.
\end{abstract}
\end{frontmatter}

\section{Introduction}\label{sec:intro}
Since early times, dealing with causality has been---and under many respects, still is---a true challenge for scientists and philosophers \citep{hume}. Nowadays, causality represents an important direction for data science, with many applications to machine learning (e.g., \citealt{scholkopf2022causality}), reinforcement learning (e.g., \citealt{zhang2020dtr}) and explainable AI (e.g., \citealt{galhotra2021explaining}).

Structural causal models are a natural formalism for causal modelling and inference, in particular for their appealing graphical representation \citep{pearl2009causality}. They are also very general and equivalent to the prominent alternative formalisms proposed to handle causality (see \citealt{ibeling2023comparing}, for instance, for a discussion on their relation with the Neyman-Rubin potential-outcome framework).

We focus in particular on Pearl's non-parametric structural causal models with discrete variables. We start by showing that they can be represented by \emph{credal networks} \citep{cozman2000credal}, which are a class of imprecise-probabilistic graphical models originally proposed as tools for sensitivity analysis in Bayesian networks. The representation is exact: every query in the causal model can be reformulated as a query in the credal network and solved by standard algorithms for the latter. The output is made of a lower and an upper bound in the case of partially identifiable queries, this typically being the case for counterfactuals, or a sharp value for identifiable ones such as the interventional queries we might compute by Pearl's do calculus. Yet, credal network inference remains a challenging task: exact inference belongs to a complexity class higher than that of Bayesian networks and the existing approximate schemes can in practice be directly applied only to a specific class of structural models. In fact, we use the relation between structural causal models and credal networks to prove that inference on the former is NP-hard even when the causal graph is a polytree.

To bypass such limitations, we derive an \emph{expectation-maximisation} (EM) scheme that reduces the credal network inference to an iterated EM with multiple initialisations. It can be understood as a sampling approach that yields a set of points inside the exact counterfactual bounds, via inferences on Bayesian networks with the same topology of the original model. Lower and upper values of these points define an approximate \emph{range} of values for the counterfactual that is encompassed by the actual bounds (we say it is an \emph{inner approximation}). We derive credibility intervals to evaluate the quality of such a range in the form of distance from the exact bounds.

As a matter of fact, relatively few works have considered so far the case of partially identifiable problems, for which bounds need to be computed. One of the first attempts in this direction is from \citet{balke1997bounds}, where the computation of bounds on interventional queries is reduced to a linear program. This allows to compute tight bounds, but the size of the program grows exponentially large, being in practice feasible only for very small models. 
\citet{kang2012inequality} present a more systematic technique to derive constraints on the query, but still with exponential growth and no explicit methods to compute bounds. \citet{sachs2020} detect a class of models allowing for an efficient reduction to linear programs, but no guarantees are provided in the general case.
More recently, both \citet{zhang2021} and \cite{duarte2021} consider an exact reduction to polynomial programming, with the former giving in addition an approximate sampling scheme.

The paper is organised as follows: after providing background material in Section~\ref{sec:background}, we discuss the notion of identifiability for causal queries, as well as that of M-compatibility, i.e., the logical consistency between a structural causal model and an empirical distribution, in Section~\ref{sec:id}. An algorithm to convert a causal model into a credal network, whose quantification is defined by the observational data, is provided in Section~\ref{sec:cn}. The approximate EM solution is instead reported in Section~\ref{sec:emcc}. Experiments on synthetic and real data are discussed in Section~\ref{sec:experiments} along with a case study. Conclusions are presented in Section~\ref{sec:conc}, proofs are gathered in~\ref{app:proofs}, while \ref{app:converg} contains additional technical results. Finally, \ref{app:faq} provides a list of questions and answers intended to further clarify various aspects of the paper.\footnote{The results we present here as a journal contribution are a revised and extended version, with new experiments on synthetic and real data and a case study, of material originally presented in three conference papers. The credal network mapping has been introduced by \citet{zaffalon2020}, the EM scheme by \citet{zaffalon2021}, and the general formulation for the credibility intervals by \citet{zaffalon2022}. The main focus of the last paper is the analysis of biased data, which is not included here.}

\section{Background}\label{sec:background}
In this section we review the necessary background on probabilistic graphical models, namely Bayesian and credal networks (Section~\ref{sec:bn_cn}) and structural causal models (Section~\ref{sec:scm}), while also discussing typical inference tasks on all these models (Section~\ref{sec:inf}). For more extensive discussions on these topics we point the reader to the books of \citet{koller2009} and \citet{pearl2009causality}. 

\subsection{Bayesian and Credal Networks}\label{sec:bn_cn}
Variable $X$ is assumed to take values from the set $\Omega_X$. The generic element of $\Omega_X$ is denoted as $x$. Here we only consider discrete variables, i.e., $|\Omega_X|<+\infty$. Denote as $P(X)$ a probability mass function (PMF) over $X$, and as $K(X)$ a \emph{credal} set (CS), which is a set of PMFs over $X$. Given variables $X$ and $Y$, a conditional probability table (CPT) $P(X|Y)$ is a collection of (conditional) PMFs indexed by the values of $Y$, i.e., $\{P(X|y)\}_{y\in\Omega_Y}$. If all PMFs in a CPT are \emph{degenerate}, i.e., there is a state receiving probability mass one and hence all the other ones receive zero, we say that also the CPT is degenerate. A credal CPT (CCPT) $K(X|Y)$ is similarly a collection of CSs over $X$ indexed by the values of $Y$. With a small abuse of terminology, we might call CPT (CCPT) also a single PMF (CS).

Consider a joint variable $\bm{X}:=(X_1,\ldots,X_n)$ and a directed acyclic graph $\mathcal{G}$ whose nodes are in a one-to-one correspondence with the variables in $\bm{X}$. Note that we use a node in $\mathcal{G}$ and its corresponding variable interchangeably. Given $\mathcal{G}$, a Bayesian network (BN) is a collection of CPTs  $\{P(X_i|\mathrm{Pa}_{X_i})\}_{i=1}^n$, where $\mathrm{Pa}_{X_i}$ denotes the \emph{parents} of $X_i$, i.e., the direct predecessors of $X_i$ according to $\mathcal{G}$. A BN induces a joint PMF $P(\bm{X})$ that factorises as follows: 
\begin{equation}\label{eq:bn}
P(\bm{x})=\prod_{i=1}^n P(x_i|\mathrm{pa}_{X_i})\,,    
\end{equation}
for each $\bm{x}\in\Omega_{\bm{X}}$, where $(x_i,\mathrm{pa}_{X_i})\sim \bm{x}$, which we use to denote that $x_i$ and $\mathrm{pa}_{X_i}$ are the values of $X_i$ and $\mathrm{Pa}_{X_i}$ consistent with $\bm{x}$ for each $i=1,\ldots,n$. A credal network (CN) is similarly intended as a collection of CCPTs. A CN defines a joint CS $K(\bm{X})$ whose elements are PMFs factorising as those of a BN with CPT values taken from the corresponding CCPTs. Some authors also require CS convexity, but this is irrelevant for the inferences we consider in this paper.

\subsection{Structural Causal Models}\label{sec:scm}
Let us first define a \emph{structural equation} (SE) $f_Y$ associated with variable $Y$ and based on the input variable(s) $X$ as a surjective function $f_Y:\Omega_{X} \to \Omega_Y$ that determines the value of $Y$ from that of $X$. We shall represent such a SE via a degenerate CPT $P(Y|X)$ such that $P(y|x):=\llbracket f_Y(x)=y \rrbracket$ for each $y\in\Omega_Y$ and $x\in\Omega_X$, where $\llbracket \cdot \rrbracket$ denote the Iverson brackets that take the value one if the statement inside the brackets is true and zero otherwise.

Consider two sets of variables $\bm{U}$ and $\bm{V}$, to be called, respectively, \emph{exogenous} and \emph{endogenous}. A collection of SEs $\{f_V\}_{V\in\bm{V}}$ such that the input variables of $f_V$ are in $(\bm{U},\bm{V})$ for each $V\in\bm{V}$ is called a \emph{partially specified structural causal model} (PSCM) over $(\bm{U},\bm{V})$. It coincides with the notion of a `functional causal model' in \citet[Chapter~1.4]{pearl2009causality}.

A PSCM $M$ induces the specification of a directed, so-called \emph{causal}, graph $\mathcal{G}$ whose nodes are in a one-to-one correspondence with the variables in $(\bm{U},\bm{V})$ and such that there is an arc between two variables if and only if the first variable is an input variable for the SE of the second. The exogenous variables are therefore root nodes of $\mathcal{G}$. We focus on \emph{semi-Markovian} PSCMs, i.e., such that their graph is acyclic. Moreover, if there is no exogenous variable with more than one endogenous child, we call the PSCM \emph{Markovian}.

In a (semi-Markovian) PSCM $M$, a joint state of $\bm{V}$ is obtained from a (joint) state of $\bm{U}$ by applying the SEs of $M$ consistently with a topological order for $\mathcal{G}$. A \emph{fully specified structural causal model} (FSCM, see \citeauthor{pearl2009causality}, \citeyear{pearl2009causality}, top of p.~69) is just a PSCM $M$ paired with a collection of marginal PMFs, one for each exogenous variable. As SEs induce (degenerate) CPTs, overall, an FSCM provides a BN specification based on $\mathcal{G}$ whose joint PMF factorises according to Equation~\eqref{eq:bn}, i.e.:
\begin{equation}\label{eq:fulljoint}
P(\bm{u},\bm{v})= \prod_{U\in\bm{U}} {\color{myblueR2}{P(u)}} \cdot \prod_{V\in\bm{V}} P(v|\mathrm{pa}_V) \,,
\end{equation}
where, for each $\bm{u}\in\Omega_{\bm{U}}$ and $\bm{v}\in\Omega_{\bm{V}}$, $(u,v,\mathrm{pa}_V)\sim(\bm{u},\bm{v})$, $\mathrm{Pa}_V$ are the parents of $V$ according to $\mathcal{G}$ (i.e., the inputs of SE $f_V$).

\begin{figure}[htp!]
\centering
\begin{tikzpicture}[scale=1.]
\node[dot,label=above left:{$V_1$}] (x1)  at (0,0) {};
\node[dot2,label=left:{$U_1$}] (u1)  at (0,-1) {};
\node[dot2,label=left:{$U_2$}] (u2)  at (2,-1) {};
\node[dot2,label=above left:{$U_3$}] (u3)  at (2,1) {};
\node[dot,label=above left:{$V_2$}] (x2)  at (1,0) {};
\node[dot,label=above left:{$V_3$}] (x3)  at (2,0) {};
\node[dot,label=above left:{$V_4$}] (x4)  at (3,0) {};
\draw[a] (u1) -- (x1);
\draw[a] (u3) -- (x3);
\draw[a] (u2) -- (x2);
\draw[a] (u2) -- (x4);
\draw[a2] (x1) -- (x2);
\draw[a2] (x2) -- (x3);
\draw[a2] (x3) -- (x4);
\end{tikzpicture}
\caption{The causal graph of a semi-Markovian PSCM.}
\label{fig:pscm}
\end{figure}
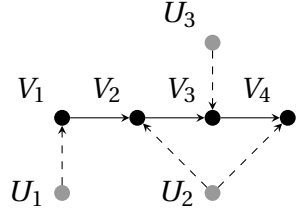

Let us clarify the above concepts by means of a small example.
\begin{example}\label{ex:pscm}
Given the endogenous variables $\bm{V}:=(V_1,V_2,V_3,V_4)$ and the exogenous ones $\bm{U}:=(U_1,U_2,U_3)$, SEs $f_{V_1}(U_1)$, $f_{V_2}(U_2,V_1)$, $f_{V_3}(U_3,V_2)$, and $f_{V_4}(U_2,V_3)$ define a semi-Markovian PSCM based on the causal graph in Figure~\ref{fig:pscm}. An FSCM based on the same SEs induces a joint PMF that factorises as in Equation~\eqref{eq:fulljoint}. By expressing the (degenerate) probabilities in the endogenous CPTs through the SEs we can therefore write the joint probability as:
\begin{equation}
\begin{split}
&P(u_1,u_2, u_3,v_1,v_2,v_3,v_4)=
{\color{myblueR2} P(u_1)P(u_2)P(u_3)}\\ 
& \cdot \llbracket f_{V_1}(u_1) = v_1 \rrbracket \llbracket f_{V_2}(u_2,v_1) = v_2 \rrbracket  \llbracket f_{V_3}(u_3,v_2) = v_3 \rrbracket \llbracket f_{V_4}(u_2,v_3) = v_4 \rrbracket\,,
\end{split}
\end{equation}
for each $v_i \in \Omega_{V_i}$ and $u_j \in \Omega_{U_j}$ with $i=1,2$ and $j=1,2,3,4$.
\end{example}

Given the causal graph $\mathcal{G}$ of a PSCM (or FSCM) $M$, obtain $\mathcal{G}'$ by removing from $\mathcal{G}$ any arc connecting pairs of endogenous variables. Let $\{\mathcal{G}_c\}_{c\in\mathcal{C}}$ denote the connected components of $\mathcal{G}'$ with $\mathcal{C}=\{1,\ldots,n_C\}$ and $n_C$ being the number of connected components of $\mathcal{G}'$. The \emph{c-components} of $M$ are the elements of the partition $\{\bm{V}^{c}\}_{c\in\mathcal{C}}$ of $\bm{V}$, where $\bm{V}^{c}$ denotes the endogenous nodes in $\mathcal{G}_c$, for each $c\in\mathcal{C}$ \citep{tian2002studies}. This procedure also induces a partition of $\bm{U}$, similarly denoted as $\{\bm{U}^{c}\}_{c\in\mathcal{C}}$. Moreover, for each $c\in\mathcal{C}$, let $\bm{W}^{c}$ denote the union of the endogenous parents of the nodes in $\bm{V}^{c}$ and $\bm{V}^{c}$ itself. Finally, for each $V\in \bm{V}^{c}$, obtain $\bm{W}_V$ by removing from $\bm{W}^{c}$ the nodes topologically following $V$ and $V$ itself (we dropped the superscript $c$ as this can be implicitly retrieved from $V$).

\citet{tian2002studies} shows that the joint PMF $P(\bm{V})$ obtained by marginalising the exogenous variables out of the joint PMF in Equation~\eqref{eq:fulljoint} is a BN that factorises as follows:
\begin{equation}\label{eq:empirical}
P(\bm{v})=\prod_{V\in\bm{V}} P(v|\bm{w}_V)\,,
\end{equation}
for each $\bm{v}\in\Omega_{\bm{V}}$ with $(v,\bm{w}_V)\sim \bm{v}$ and $V\in\bm{V}$ (remember that symbol $\sim$ is used to denote consistent instances of variables as from Section~\ref{sec:bn_cn}). In the following we call such a BN the \emph{endogenous} BN of a FSCM. 

For Markovian models, where the partitions induced by the c-components are made of singletons, we trivially have $\bm{W}_V=\mathrm{Pa}_V \cap \bm{V}$ for each $V\in\bm{V}$, i.e., the parents of an endogenous node in the endogenous BN are its endogenous parents in the original model. For non-Markovian models, we illustrate the procedure by the following example.

\begin{figure}[htp!]
\centering
\begin{tikzpicture}[scale=0.9]
\node[dot,label=above left:{$V_1$}] (x1)  at (0,0) {};
\node[dot,label=above left:{$V_2$}] (x2)  at (1,0) {};
\node[dot,label=above left:{$V_3$}] (x3)  at (2,0) {};
\node[dot,label=above left:{$V_4$}] (x4)  at (3,0) {};
\draw[a2] (x1) -- (x2);
\draw[a2] (x2) -- (x3);
\draw[a2] (x3) -- (x4);
\draw[a2] (x1) [out=-40,in=-130] to (x4);
\draw[a2] (x2) [out=-40,in=-150] to (x4);
\end{tikzpicture}
\caption{The graph of the endogenous BN for the FSCM in Figure \ref{fig:pscm}.}
\label{fig:ebn}
\end{figure}
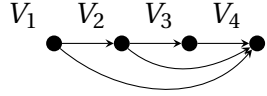

\begin{example}\label{ex:ccom}
The FSCM in Example \ref{ex:pscm} has three c-components inducing the endogenous partition $\{ \bm{V}^c\}_{c=1}^3$ with $\bm{V}^{1}:=\{V_1\}$, $\bm{V}^{2}:=\{V_2,V_4\}$, and $\bm{V}^{3}:=\{V_3\}$, and the exogenous partition $\{\bm{U}^c\}_{c=1}^3$ with $\bm{U}^{1}:=\{U_1\}$, $\bm{U}^{2}:=\{U_2\}$, and $\bm{U}^{3}:=\{U_3\}$. We easily obtain 
$\bm{W}^1=\{V_1\}$, $\bm{W}^2=\{V_1,V_2,V_3,V_4\}$, and $\bm{W}^3=\{V_2,V_3\}$. Finally, by considering the unique topological order over the four endogenous variables, we get $\bm{W}_{V_1}=\emptyset$, $\bm{W}_{V_2}=\{V_1\}$, $\bm{W}_{V_3}=\{V_2\}$, and $\bm{W}_{V_4}=\{V_1,V_2,V_3\}$. The endogenous BN obeying the factorisation in Equation~\eqref{eq:empirical} therefore corresponds to the graph in Figure~\ref{fig:ebn}.
\end{example}

In practice, the CPTs in the right-hand side of Equation~\eqref{eq:empirical} can be computed through standard BN inference algorithms by just regarding the FSCM as a BN. With PSCMs, assuming the availability of a dataset $\mathcal{D}$ of endogenous observations, we might also quantify the endogenous BN by learning the probabilities in the CPTs from $\mathcal{D}$.

Finally, let us note that PSCMs assume SEs to be given. However, in the case where endogenous variables take on finitely many values, recent results allow one to specify a PSCM by only giving a causal graph; SEs are automatically defined, without loss of generality (yet adding a likely excess of caution to the model), via a \emph{canonical specification}. Let us introduce it in the simple case of Markovian models: we say that SE $f_V$ is canonical if the states of the (single because of Markovianity) exogenous parent $U$ of $V$ index all the  deterministic relations between the endogenous parents (i.e., $\bm{W}_V$ because of Markovianity) and $V$. This requires:
\begin{equation}\label{eq:canonical}
|\Omega_U|=|\Omega_V|^{\prod_{W\in\bm{W}_V} |\Omega_W|}\,.
\end{equation}
A Markovian PSCM whose SEs are all canonical is also called canonical. We refer the reader to the works of \citet{duarte2021} and \citet{zhang2021} for a generalisation of such a concept to non-Markovian models.

\subsection{Updating, Interventions and Counterfactuals}\label{sec:inf}
Computing the posterior probability for the state of a queried variable given an evidence with respect to the joint PMF of a BN, as well as the bounds of this probabilities with respect to the joint CS of a CN are NP-hard tasks (e.g., \citeauthor{maua2014probabilistic}, \citeyear{maua2014probabilistic}). Yet, polynomial algorithms computing approximate inferences for the general case (e.g., \citeauthor{approxlp}, \citeyear{approxlp}) or exact ones for classes of sub-models (e.g., \citeauthor{2u}, \citeyear{2u}) are available for CNs, not to mention the copious tools for BNs (e.g., \citeauthor{koller2009}, \citeyear{koller2009}).

Observational queries in FSCMs can be addressed in the endogenous BN, and the same can be done for PSCMs by assuming the availability of the dataset $\mathcal{D}$ of endogenous observations. 

To perform causal inference, \emph{interventions} denoted as $\mathrm{do}(\cdot)$ should be considered instead. In an FSCM or PSCM $M$, given $V \in \bm{V}$ and $v\in\Omega_V$, $\mathrm{do}(V=v)$ simulates a physical action on $M$ forcing $V$ to take the value $v$. The original SE $f_V$ should be consequently replaced by a constant map $V=v$. Notation $M_v$ is used for such a modified model, whose graph is obtained by removing from $\mathcal{G}$ the arcs entering $V$, and for which evidence $V=v$ is considered. In an FSCM $M$, given $V,W\in\bm{V}$ and $v\in\Omega_V$, $P(w|\mathrm{do}(v))$ denotes the conditional probability of $W=w$ in the post-intervention model, i.e., $P'(w|v)$, where $P'$ is the joint PMF induced by $M_v$. As interventions commute, there are no ordering issues when coping with multiple interventions. If evidence is also available, i.e., some variables have been observed, it is customary to assume that observations take place after the interventions. Note that interventional queries assume the set of observed variables and that of intervened variables disjoint. 

A more general setup is provided by \emph{counterfactual} queries, where the same variable may be observed as well as subject to intervention, albeit in distinct `worlds'. In mathematical parlance, if $\bm{W}$ are the queried variables, $\bm{V}'$ the observed ones and $\bm{V}''$ the intervened ones, we write the query by $P(\bm{W}_{\bm{v}''}|\bm{v}')$ with possibly $\bm{V}'\cap \bm{V}''\neq \emptyset$. A popular counterfactual query involving two endogenous Boolean variables $X$ and $Y$ of an FSCM is the \emph{probability of necessity} (PN), i.e., the probability that event $Y$ would not have occurred by disabling $X$, given that $X$ and $Y$ did in fact occur. This corresponds to $P(Y_{X=0}=0|X=1,Y=1)$. Similarly, the probability of \emph{sufficiency} (PS) is the probability that $Y$ would have occurred by activating $X$, given that $X$ and $Y$ did not occur, i.e., $P(Y_{X=1}=1|X=0,Y=0)$. Finally, the probability of \emph{necessity and sufficiency} (PNS), corresponding to $P(Y_{X=1}=1,Y_{X=0}=0)$, is the probability that $Y$ would respond to $X$ both ways, thus measuring both the sufficiency and necessity of $X$ to produce $Y$. A characterisation of these three counterfactual probabilities has been provided by \citet{pearl1999probabilities}.

\begin{figure}[htp!]
\centering
\begin{tikzpicture}[scale=1.0]
\node[dot2,label=left:{$U_1$}] (u1)  at (0,-1) {};
\node[dot2,label=left:{$U_2$}] (u2)  at (2,-1) {};
\node[dot2,label=above left:{$U_3$}] (u3)  at (2,1) {};
\node[dot,label=above left:{$V_1$}] (x1)  at (0,0) {};
\node[dot,label=above left:{$V_2$}] (x2)  at (1,0) {};
\node[dot,label=above left:{$V_3$}] (x3)  at (2,0) {};
\node[dot,label=above left:{$V_4$}] (x4)  at (3,0) {};
\node[dot,label=below left:{$V_1'$}] (y1)  at (0,-2) {};
\node[dot,label=below left:{$V_2'$}] (y2)  at (1,-2) {};
\node[dot,label=below left:{$V_3'$}] (y3)  at (2,-2) {};
\node[dot,label=below left:{$V_4'$}] (y4)  at (3,-2) {};
\draw[a] (u1) -- (x1);
\draw[a] (u1) -- (y1);
\draw[a] (u3) -- (x3);
\draw[a] (u3) edge[bend left=30] node[left] {} (y3); 
\draw[a] (u2) -- (x2);
\draw[a] (u2) -- (x4);
\draw[a] (u2) -- (y2);
\draw[a] (u2) -- (y4);
\draw[a2] (x1) -- (x2);
\draw[a2] (x2) -- (x3);
\draw[a2] (x3) -- (x4);
\draw[a2] (y1) -- (y2);
\draw[a2] (y2) -- (y3);
\draw[a2] (y3) -- (y4);
\end{tikzpicture}
\caption{The graph of the twin network for the model in Figure \ref{fig:pscm}.}\label{fig:twin}
\end{figure}
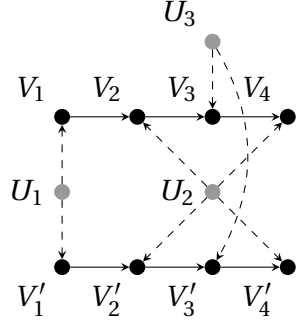

Computing counterfactual queries in an FSCM may be achieved via an auxiliary structure called a \emph{twin network}  \citep{balke1994counterfactual}. This is simply an FSCM where the original endogenous nodes (and their SEs) have been duplicated, while remaining affected by the same exogenous variables. As an example, Figure~\ref{fig:twin} depicts the twin network for the model in Figure~\ref{fig:pscm}. More general (e.g., involving more than two copies of the same endogenous node) and compact structures can be also considered \citep{shpitser2007counterfactuals}. Computing a counterfactual in the twin network of an FSCM is analogous to what is done with interventional queries provided that interventions and observations are associated with distinct copies of the same variable. BN inference eventually allows one to compute the counterfactual query in such an augmented model. 

In the next section we discuss the computation of causal queries in PSCMs.

\section{Identifiability and M-Compatibility}\label{sec:id}
As shown in the previous section, FSCMs allow to compute any causal query by standard BN inference algorithms. Yet, as the exogenous variables are typically latent, the marginal PMFs over them are rarely available and we should instead cope with a PSCM (i.e., the SEs) only. We assume the PSCM M comes paired up with a dataset $\mathcal{D}$ of endogenous observations. 

{\color{myblueR2} In the following we first formalise the concept of identifiability and discuss how to address partially identifiable queries (Section~\ref{sec:identifiability}). This is based on the central notion of M-compatibility and its likelihood-based reformulation that becomes our main operational tool 
(Section~\ref{sec:compatiblity}).}

{\color{myblueR2}\subsection{Partial Identifiability and M-Compatibility}\label{sec:identifiability}
As noticed in Section \ref{sec:scm}, given a PSCM, the endogenous observations in $\mathcal{D}$ can be used to quantify the CPTs of the related endogenous BN  \citep{tian2002studies}. In this paper we use maximum likelihood estimation, whence we set:
\begin{equation}\label{eq:freqs}
P(v|\bm{w}_V):=\frac{n(v,\bm{w}_V)}{n(\bm{w}_V)}\,,
\end{equation}
for each $v\in\Omega_V$, $\bm{w}_V\in\Omega_{\bm{W}_V}$, and $V\in\bm{V}$, with $n(\cdot)$ denoting the frequencies in $\mathcal{D}$ of its argument.

We take the endogenous BN obtained in this way as our \emph{ground truth}: it represents the joint mass function $P(\bm{V})$ on the observable variables that results from the SCM we are given in input and the available data. Of course, this is not the same BN that we would obtain in the limit of infinite data; let us remark that we are not concerned with limit considerations in this paper, and we rather work with finite samples and the related estimates. Consequently, $P(\bm{V})$ is taken as our `objective' piece of information on the domain, which is not under question. $P(\bm{V})$ is not sufficient to draw causal inference, though. For this we need to go back to our SCM.

But the question is that there are in general many FSCMs that lead to the very same $P(\bm{V})$. Using Equation~\eqref{eq:fulljoint}, we can characterise them as follows:
\begin{equation}\label{eq:ident}
\sum_{\bm{u}\in\Omega_{\bm{U}}} \left[ \prod_{U\in\bm{U}} P(u) \cdot \prod_{V\in\bm{V}} P(v|\mathrm{pa}_V) \right] = P(\bm{v})\,,
\end{equation}
for each $\bm{v}\in\Omega_{\bm{V}}$, with $(v,\mathrm{pa}_V)\sim \bm{v}$ and $u\sim\bm{u}$. 
We call this \emph{M-compatibility} to emphasise that it implicitly defines all and only the models (FSCMs) that are compatible with the PSCM given in input as well as with the data, in the sense that they all yield $P(\bm{V})$; we denote the set of all these models by $\mathcal{M}_{M,P}$.

Let us illustrate this multiplicity with a simple example.}


\begin{example}\label{ex:single}
Consider a PSCM with a Boolean endogenous variable $V$, a ternary exogenous parent $U$ with $\Omega_U:=\{0,1,2\}$, and a SE such that $f_V(U=0)=0$ and $f_V(U=1)=f_V(U=2)=1$. A dataset $\mathcal{D}$ of observations for $V$ allows to assess the endogenous PMF
$P(V)$. Equation~\eqref{eq:ident} rewrites therefore as:
\begin{equation}
\sum_{u=0}^2 P(u) \cdot P(v|u) = P(v)\,,
\end{equation}
for $v=0,1$. This is a linear system with solution:
\begin{equation}
P(U)=\left[
\begin{array}{c}
P(V=0) \\ \alpha \\ P(V=1)-\alpha
\end{array}
\right]\,,
\end{equation}
for each $\alpha \in [0,1-P(V=0)]$.
\end{example}

{\color{myblueR2} That $\mathcal{M}_{M,P}$ is not a singleton in general is the reason why one usually talks of `partial identifiability' when it comes to causal inference: namely, that it is not possible in general to identify a single causal model that is consistent with the data; the best one can do is to rather consider the set $\mathcal{M}_{M,P}$. From this it also follows the notion of `partially identifiable query': this is a causal query, such as an intervention or a counterfactual, whose outcome is partially indeterminate due to the multiplicity of FSCMs in $\mathcal{M}_{M,P}$ we have to query to get a answer. If we ask for an expectation, for instance, each FSCM $M'$ will yield a number $q_{M'}$, and all these numbers will eventually be summarised by a lower and an upper value, namely the shortest interval that contains all the delivered expectations:
\begin{equation}\label{eq:bounds}
\left[\min_{M' \in \mathcal{M}_{M,P}} q_{M'}, \max_{M' \in \mathcal{M}_{M,P}} q_{M'}\right].
\end{equation}
Sometimes the interval will naturally collapse to a number, thus embodying the case of an identifiable query. But this will just be a special case. It occurs, for example, when the \emph{do calculus} of \cite{pearl2009causality} and its extensions (see, e.g., \citeauthor{bareinboim2012causal}, \citeyear{bareinboim2012causal}) can reduce interventional queries on PSCMs to observational ones (since the do calculus is sound and complete, it can be used to have an equivalent condition to identifiability in the interventional case).

Our focus in the rest of the paper is however on partial identifiability, which in practice means the computation of the bounds in Equation~\eqref{eq:bounds}.}

\subsection{M-Compatibility and Likelihood Maximisation}\label{sec:compatiblity}
{\color{myblueR2} To address causal queries by Equation~\eqref{eq:bounds},
we need the set $\mathcal{M}_{M,P}$ to be non-empty, this meaning that Equation~\eqref{eq:ident} should admit at least one FSCM solution. If this was not the case, the endogenous distribution $P(\bm{V})$ would be incompatible with the PSCM M under consideration, in the sense that no FSCM based on the same SEs of the PSCM could generate such a distribution when the latent variables get marginalised out.

Remember that M-compatibility is about the possibility of reconstructing the uncertainty about $\bm{U}$ from the PSCM together with the endogenous PMF $P(\bm{V})$. If it fails, we know that the task is hopeless, because there is no $P(\bm{U})$ that can eventually lead to $P(\bm{V})$. This may happen either because the sample is too small, or because the PSCM is a wrong model of the phenomenon under study. In either case, one should refrain from making inferences using jointly the PSCM and $P(\bm{V})$ as they would be logically contradicting each other.

The notion of M-compatibility is central to this paper. Now we proceed to reformulate it in a way that makes it easier to use it as an operational tool.}

By exploiting the factorisation in Equation~\eqref{eq:fulljoint}, while also grouping together the terms corresponding to the different c-components, the log-likelihood of $\mathcal{D}$ from an FSCM can be written as follows:
\begin{equation}\label{eq:liku}
l(\theta_{\bm{U}}):=\sum_{c\in\mathcal{C}} \sum_{\bm{w}^{c}\in\Omega_{\bm{W}_C}} n(\bm{w}^{c}) \cdot \log \sum_{\bm{u}^{c}\in\Omega_{\bm{U}_C}} \left[ \prod_{U\in\bm{U}^{c}} \theta_u \cdot \prod_{V\in\bm{V}^{c}} P(v|\mathrm{pa}_V) \right]
\,,
\end{equation}
with $(u,v,\mathrm{pa}_V)\sim (\bm{u}^c,\bm{w}^c)$ for each $c\in\mathcal{C}$ and $\theta_{\bm{U}}:=( \theta_u)_{\substack{U \in \bm{U}, u \in \Omega_U}}$, {\color{myblueR2}with $\theta_u$ being the unknown `true' chance of $U=u$}. For the endogenous BN of an FSCM, the log-likelihood of $\mathcal{D}$ is instead:
\begin{equation}\label{eq:likdec}
\lambda(\theta_{\bm{V}}):=\sum_{V\in\bm{V}} \sum_{v,\bm{w}_V} n(v,\bm{w}_V) \cdot \log \theta_{v|\bm{w}_V}\,, 
\end{equation}
where, for the sake of a light notation, the domains of the sums over $v$ and $\bm{w}_V$ are left implicit and $\theta_{\bm{V}}:=(\theta_{v|\bm{w}_V})_{v \in \Omega_V,\bm{w}_V\in\Omega_{\bm{W}_V},V\in\bm{V}}$. 
Note that the conditional chances in $\theta_{\bm{V}}$ can be directly obtained from those in $\theta_{\bm{U}}$.

Equation~\eqref{eq:likdec} exhibits the decomposable structure of a multinomial likelihood. Such a concave function has a unique global maximum achieved where the relative frequencies in Equation~\eqref{eq:freqs} are attained by the arguments $\theta_{\bm{V}}$, and no local maxima (see, e.g., \citeauthor{koller2009}, \citeyear{koller2009}). Let us denote as $\lambda^*$ the value of global maximum of the function in Equation~\eqref{eq:likdec}.

\begin{theorem}\label{th:unimod}
Let $\mathcal{K}$ denote the set of joint mass functions $P(\bm{U})$ consistent with Equation~\eqref{eq:ident}. For $\mathcal{K} \neq \emptyset$, the log-likelihood $l$ in Equation~\eqref{eq:liku} achieves its global maximum, equal to $\lambda^*$, if and only if $\theta_{\bm{U}}\in\mathcal{K}$. For $\mathcal{K}=\emptyset$, instead, $l$ takes only values strictly smaller than $\lambda^*$.
\end{theorem}
In other words, global optimality of the log-likelihood is tantamount to finding an FSCM based on the given PSCM compatible with the data, this also allowing to decide M-compatibility, as formalised by the following result.
\begin{corollary}\label{cor:k}
The function in Equation~\eqref{eq:liku} achieving its global maximum $\lambda^*$ is an equivalent condition for M-compatibility.
\end{corollary}

{\color{myblueR2}This corollary is important because it allows us to narrow the attention to the M-compatible models only via likelihoods, thus enabling us to use a simple numerical test of M-compatibility.

Before using such a test, in the next section we show how CNs can be used to address the exact computation of a partially identifiable query as well as the test of M-compatibility in a way alternative to the one provided by Corollary~\ref{cor:k}.}

\section{Credal Networks for Partial Identifiability}\label{sec:cn}
CNs offer a suitable formalism to address partial identifiability as in Equation~\eqref{eq:bounds}. Exactly as a single FSCM is equivalent to a BN, the FSCMs in $\mathcal{M}_{M,P}$ (all with the same SEs) can be regarded as a CN. To achieve that, a trivial but highly inefficient approach, inspired by \citet{antonucci2008b}, would consist in adding an auxiliary parent variable whose states index all the exogenous PMF specifications in $\mathcal{K}$, these clearly being in a one-to-one correspondence with the elements of $\mathcal{M}_{M,P}$. We derive instead a compact CN specification for Markovian models (Section \ref{sec:markov}), and then extend it to more general cases (Section \ref{sec:quasimarkov}). Such a mapping allows to virtually reduce causal inference to CN inference, while also leading to a new characterisation of the computational complexity of the latter. Those results are demonstrated by a number of examples leading to a deeper discussion on the importance of testing M-compatibility (Section~\ref{sec:incomp}).

\subsection{Credal Network Mapping in the Markovian Case}\label{sec:markov}
We already noticed how the process of specifying the endogenous BN for Markovian PSCMs is particularly  simple: $\bm{W}_V$, i.e., the parents of $V$ in the endogenous BN, are just the endogenous parents of $V$ in the original model, for each $V\in\bm{V}$. This allows for a decomposition of the non-linear constraints in Equation~\eqref{eq:ident}, which are mixing the marginal PMFs for the different exogenous variables. We consequently obtain a separate set of linear constraints for each exogenous variable. This corresponds to a CS specification for each $U \in \bm{U}$ and, overall, a proper CN with no need of auxiliary variables. The procedure is shown in Algorithm~\ref{alg:simple}, where, just for the sake of readability, we ignore the case of endogenous nodes without exogenous parents and we do not explicitly write the normalisation and non-negativity constraints of CSs.

\begin{algorithm}[htp!]
{\color{myblueR2}\caption{CN map for the Markovian case.}\label{alg:simple}
\SetKwInOut{Input}{input}\SetKwInOut{Output}{output}
\Input{A Markovian PSCM M over $(\bm{U},\bm{V})$ and a BN $P(\mathbf{V})$}
\Output{A credal set $K(\mathbf{U})$}
\For {$V\in\bm{V}$}{
$U \leftarrow \mathrm{Pa}_V \cap \bm{U}$\;
$\bm{W}_V \leftarrow \mathrm{Pa}_V \setminus \{ U \}$\;
$K(U) \leftarrow \left\{ P(U) \left|
\begin{array}{l}
\sum_{u \in \Omega_U} P(u) \cdot \llbracket f_V(u,\bm{w}_V)=v \rrbracket = P(v|\bm{w}_V)\\
\forall v \in \Omega_V ,\bm{w}_V \in \Omega_{\bm{W}_V}
\end{array}\right.\right\}$
}}
\end{algorithm}
{\color{myblueR2}Note that we try to keep notation simple by denoting the collection of credal sets $\{K(U)\}$, with $U\in\bm{U}$, $K(\bm{U})$---while also abusing terminology by calling the latter a credal set.}

The algorithm correctness is guaranteed by the following result. 

\begin{theorem}\label{th:markov}
For a Markovian PSCM M, the FSCMs of $\mathcal{M}_{M,P}$ can be represented as the BNs of a CN whose CCPTs are those induced by the SEs of M and the CS $K(\mathbf{U})$ returned by Algorithm~\ref{alg:simple}.
\end{theorem}

The result implies that in the Markovian case we can regard the computation of the bounds of a PSCM query in Equation~\eqref{eq:bounds} as an inference task on the CN based on the output of Algorithm~\ref{alg:simple}. In spite of the hardness of CN inference, the mapping is not increasing the complexity of PSCM queries as shown by the following result.

\begin{theorem}\label{th:hard}
The computation of post-interventional queries for single variables in polytree-shaped PSCMs is NP-hard.
\end{theorem}

The proof of the above result, reported in \ref{app:proofs}, is based on the analogous complexity result for CNs derived by \citet{cozman2002}. Other complexity results for CNs might be similarly applied to PSCMs by exploiting the CN formulation proposed by \citet{cozman2017}.

In the next section we discuss how to cope with non-Markovian PSCMs.

\subsection{Beyond Markovianity}\label{sec:quasimarkov}
In order to extend the procedure discussed in the previous section, let us first consider a simple non-Markovian example.

\begin{figure}[htp!]
\centering
\begin{tikzpicture}[scale=1.0]
\node[dot,label=left:{$V_1$}] (1)  at (0,0) {};
\node[dot,label=right:{$V_2$}] (2)  at (2,0) {};
\node[dot2,label=above:{$U$}] (u)  at (1,1) {};
\draw[a] (u) -- (1);
\draw[a] (u) -- (2);
\draw[a2] (1) -- (2);
\end{tikzpicture}
\caption{The causal graph of a semi-Markovian, but not Markovian, PSCM.}\label{fig:non-markov}
\end{figure}
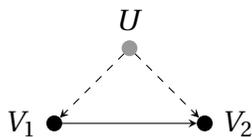

\begin{example}\label{ex:nonmarkov}
Consider a PSCM M over the graph in Figure \ref{fig:non-markov}. The endogenous variables $V_1$ and $V_2$ are Boolean, while their common exogenous parent $U$ is such that $\Omega_U:=\{0,1,\ldots,4\}$. For $V_1$, we have $f_{V_1}(U=u)$ equal to zero for $U=0,3,4$ and one for the other states of $U$. For $V_2$, we have instead $f_{V_2}(U=u,V_1=0)=0$ for $u=0,2$ and $f_{V_2}(U=u,V_1=1)=0$ for $U=2,3$, while the SEs return one in the other cases. Equation~\eqref{eq:ident} rewrites as:
\begin{equation}\label{eq:constraint33}
\sum_{u \in \Omega_U} P(u) \cdot P(v_1|u) \cdot P(v_2|u,v_1) = P(v_1,v_2)\,,
\end{equation}
to be considered for each $v_1\in\Omega_{V_1}$ and $v_2\in\Omega_{V_2}$. In the sum on the left-hand side, the terms corresponding to values of $u$ that are not simultaneously consistent, through SEs $f_{V_1}$ and $f_{V_2}$, with both $v_1$ and $v_2$, are zero. We thence rewrite Equation~\eqref{eq:constraint33} as:
\begin{equation}\label{eq:constraint3}
\sum_{u \in \Omega_U: \substack{f_{V_1}(u)=v_1,\\f_{V_2}(u,v_1)=v_2}} P(u) =P(v_1,v_2)\,.
\end{equation}
Equation~\eqref{eq:constraint3} defines a linear system analogous to the one in Example~\ref{ex:single}, whose solutions define a CS $K(U)$. As an example, if the joint endogenous PMF is such that such that 
$P(V_1=0,V_2=0)=\frac{1}{5}$,
$P(V_1=1,V_2=0)=\frac{2}{5}$, and
$P(V_1=0,V_2=1)=\frac{4}{15}$, the corresponding linear constraints becomes $P(U=0)=\frac{1}{5}$, $P(U=1)= \frac{2}{15}$ and  $P(U=2)=\frac{2}{5}$, and $P(U=3)+P(U=4)=\frac{4}{15}$. The elements of the corresponding CS $K(U)$ can be therefore parametrised as $P(U) = [\frac{1}{5},\frac{2}{15},\frac{2}{5},\alpha,\frac{4}{15}-\alpha]$ with $\alpha\in[0,\frac{4}{15}]$.
\end{example}

In the non-Markovian case, the common exogenous parents of two or more endogenous variables are called \emph{confounders}. Confounders express the lack of Markovianity also at the PSCM level, being input variables common to two or more SEs. In these cases, the surjectivity we assume for single SEs (see Section~\ref{sec:scm}) is extended to the joint SE involving all the SEs with the same confounder in input. 
For instance, the joint SE $(f_{V_1},f_{V_2})$ in Example~\ref{ex:nonmarkov} is \emph{jointly surjective} as all the four joint states of $(V_1,V_2)$ can be obtained by (at least) a value of the input variable $U$.

Let us call \emph{quasi-Markovian} a PSCM (or FSCM) such that $|\bm{U}^{c}|=1$ for each $c\in\mathcal{C}$, this meaning that no endogenous variable has more than one exogenous parent. 
This is for instance the case of the PSCM in Figure~\ref{fig:pscm}, as well as the one in Figure~\ref{fig:non-markov}. In a quasi-Markovian PSCM, notation $\mathrm{Pa}_V'$ is used for the endogenous parents of $V$, i.e., $\mathrm{Pa}_V':=\mathrm{Pa}_V \cap \bm{V}$. It is easy to check that $\mathrm{Pa}_V'\subseteq \bm{W}_V$.

A procedure analogous to that in Example~\ref{ex:nonmarkov} can be derived for any quasi-Markovian PSCM. As in the previous section, the key point is that the constraints on the marginal PMF of an exogenous variable $U\in\bm{U}$ imposed by the consistency with the endogenous PMF can be specified separately from those of the other exogenous variables. This corresponds to Algorithm \ref{alg:quasi} that allows to derive the CS $K(\mathbf{U})$ and hence obtain a CN from a quasi-Markovian model.

\begin{algorithm}[htp!]
{\color{myblueR2}\caption{CN map for the quasi-Markovian case.}\label{alg:quasi}
\SetKwInOut{Input}{input}\SetKwInOut{Output}{output}
\Input{A quasi-Markovian PSCM M over $(\bm{U},\bm{V})$ and a BN $P(\mathbf{V})$}
\Output{A credal set $K(\mathbf{U})$}
\For{$U\in\bm{U}$}{
$c \leftarrow$ index of the c-component of $U$\;
$\gamma \leftarrow \emptyset$\;
\For{$\bm{w}^{c} \in \Omega_{\bm{W}^{c}}$}{
$\Omega_U^{\bm{w}^c} \leftarrow \left\{ u \in \Omega_U \left| 
\begin{array}{l}
f_V(u,\mathrm{pa}_V')=v \\ \forall V \in \bm{V}^{c}, (v,\mathrm{pa}_V')\sim \bm{w}^{c}
\end{array}
\right. \right\}$\;
$\gamma \leftarrow \gamma \cup \left\{ \sum_{u \in \Omega_U^{\bm{w}^c}} P(u) = \prod_{V \in \bm{V}^{c}} P(v|\bm{w}_V) , (v,\bm{w}_V) \sim \bm{w}^c \right\}$
}
$K(U) \leftarrow \left\{ P(U) \left| \, \gamma \right. \right\}$}}
\end{algorithm}

It is not difficult to check that Algorithm~\ref{alg:quasi} coincides with Algorithm~\ref{alg:simple} in the case of Markovian inputs. The relation between the two procedures is even  stronger as shown by the following result, which can be regarded as a generalisation of Theorem~\ref{th:markov}.

\begin{theorem}\label{th:quasi}
For a quasi-Markovian PSCM M, the FSCMs of $\mathcal{M}_{M,P}$ are the BNs associated with a CN whose CCPTs are those induced by the SEs of M and the CS $K(\mathbf{U})$ returned by Algorithm \ref{alg:quasi}.
\end{theorem}

Regarding complexity, unlike Algorithm \ref{alg:simple}, where the number of constraints defining the CSs of the CN roughly corresponds to the size of the degenerate CPTs defining the SEs in the input PSCM, the bottleneck of Algorithm \ref{alg:quasi} is the loop over lines 4--7, which takes $O(2^{|\bm{W}^c|})$ time. For quasi-Markovian models, $\bm{W}^c$ is to the union of the children of $U$ (i.e., $\bm{V}^c$) and their parents, i.e, the Markov blanket of $U$. Setting a bound to this number (e.g., having that confounders only act on pairs of endogenous variables and a bounded indegree) would therefore make the algorithm polynomial.

The above result allows for addressing the computation of the bounds in Equation~\eqref{eq:bounds} through CN inference even in the quasi-Markovian case. In principle, any semi-Markovian model can be turned into quasi-Markovian, e.g., by clustering all $U$ variables of a c-component into a single one; yet this neglects the exponential blowup in the computation that follows as a consequence.\footnote{Note also that Theorem 2.4 of \citet{zhang2021} allows to extend our results to models with continuous exogenous variables, provided the the endogenous ones remain discrete.}

Overall we mapped a hard task (cf. Theorem~\ref{th:hard}) to another hard task \citep{maua2014probabilistic}. Compared to analogous efforts of \citet{duarte2021}, where a mapping to general polynomial programming has been derived, CN inference represents a more specific field that has been subject of intense investigation in the last three decades \citep{maua2020}, and for which dedicated and stable solvers are freely available \citep{huber2020crema}.

Moreover, as discussed in the next section, Algorithm \ref{alg:quasi} allows to exactly test the M-compatibility (see Section \ref{sec:compatiblity}) with quasi-Markovian PSCMs, this corresponding to the feasibility of the linear constraints of each $U\in\bm{U}$. In the next section we discuss this procedure by also advocating the importance of checking M-compatibility and the role of SEs to achieve that.

\subsection{M-Compatibility and the Limits of Rung 2 of Pearl's Hierarchy}\label{sec:incomp}
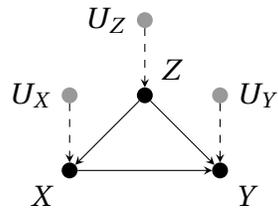
\begin{figure}[htp!]
\centering
\begin{tikzpicture}[scale=1.0]
\node[dot,label=below left:{$X$}] (x)  at (0,0) {};
\node[dot,label=below right:{$Y$}] (y)  at (2,0) {};
\node[dot,label=above right:{$Z$}] (z)  at (1,1) {};
\node[dot2,label=left:{$U_X$}] (u)  at (0,1) {};
\node[dot2,label=left:{$U_Z$}] (w)  at (1,2) {};
\node[dot2,label=right:{$U_Y$}] (v)  at (2,1) {};
\draw[a] (u) -- (x);
\draw[a] (w) -- (z);
\draw[a] (v) -- (y);
\draw[a2] (x) -- (y);
\draw[a2] (z) -- (x);
\draw[a2] (z) -- (y);
\end{tikzpicture}
\caption{The causal graph of a Markovian PSCM.\label{fig:pearl}}
\end{figure}

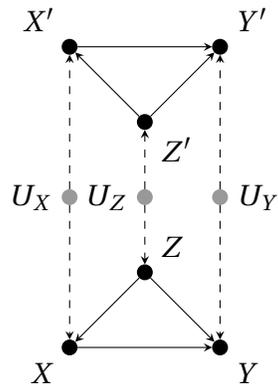
\begin{figure}[htp!]
\centering
\begin{tikzpicture}[scale=1.0]
\node[dot,label=below left:{$X$}] (x)  at (0,0) {};
\node[dot,label=below right:{$Y$}] (y)  at (2,0) {};
\node[dot,label=above right:{$Z$}] (z)  at (1,1) {};
\node[dot2,label=left:{$U_X$}] (u)  at (0,2) {};
\node[dot2,label=left:{$U_Z$}] (w)  at (1,2) {};
\node[dot2,label=right:{$U_Y$}] (v)  at (2,2) {};
\node[dot,label=above right:{$Y'$}] (y2)  at (2,4) {};
\node[dot,label=above left:{$X'$}] (x2)  at (0,4) {};
\node[dot,label=below right:{$Z'$}] (z2)  at (1,3) {};
\draw[a] (u) -- (x);
\draw[a] (w) -- (z);
\draw[a] (v) -- (y);
\draw[a2] (x) -- (y);
\draw[a] (u) -- (x2);
\draw[a] (w) -- (z2);
\draw[a] (v) -- (y2);
\draw[a2] (x2) -- (y2);
\draw[a2] (z2) -- (x2);
\draw[a2] (z2) -- (y2);
\draw[a2] (z) -- (x);
\draw[a2] (z) -- (y);
\end{tikzpicture}
\caption{The twin network of the of the PSCM in Figure~\ref{fig:pearl}.\label{fig:twinpearl}}
\end{figure}

Let us first consider a simple example to be used along the section to clarify our findings.
\begin{example}[]\label{ex:pearl}
Consider the setup from \citet[Section~4.1]{exact2021causes} referring to a study about the recovery $Y$ of patients of gender $Z$ possibly subject to treatment $X$. A sample of 700 patients is considered (the corresponding frequencies are in Table~\ref{tab:study}). The authors show that for a PSCM whose endogenous BN has the same graph of that induced from the causal graph in Figure~\ref{fig:pearl}, given the sample, the PNS for treatment ($X$) on effect ($Y$) is no greater than $0.01$ (this is the value rounded to the second decimal place; a more precise estimate is $0.015$). They tell this for any PSCM with that graph, irrespectively of the specific SEs it uses; they only require that the endogenous PMF factorises according to the (endogenous BN induced by that) graph.

Our CN approach needs instead the PSCM specification, i.e., the SEs. Let us assume M canonical. Because of Equation~\eqref{eq:canonical}, this corresponds to $|\Omega_{U_Z}|=2$, $|\Omega_{U_X}|=4$ and $|\Omega_{U_Y}|=16$. To obtain canonical SEs, we set $f_Z$ equal to the identity map, while for $f_X$:
\begin{eqnarray}
f_X(Z,U_X=0)&:=&\neg Z\,,\\f_X(Z,U_X=1)&:=&0\,,\\f_X(Z,U_X=2)&:=&Z\,,\\f_X(Z,U_X=3)&:=&1\,.
\end{eqnarray}
The relations between $Y$ and $(X,Z)$ according to $f_Y$ induced by the sixteen states of $U_Y$ can be similarly listed. M-compatibility with data in Table~\ref{tab:study} corresponds to the constraints in line 4 of Algorithm~\ref{alg:simple} (the model is Markovian). For $U_Z$, this trivially means $P(U_Z=0)=P(Z=0)=\frac{470}{700}$ and $P(U_Z=1)=P(Z=1)=\frac{230}{700}$. For $P(U_X)$ we have instead the two following independent constraints:
\begin{eqnarray}\label{eq:c1}
P(U_X=1)+P(U_X=2) &=& P(X=0|Z=0)=\frac{116}{470}\,,\\ \label{eq:c2}
P(U_X=0)+P(U_X=1) &=& P(X=0|Z=1)=\frac{120}{230}\,.
\end{eqnarray}
Four independent linear constraints can be similarly obtained for $P(U_Y)$. 

By standard linear programming tools, we might trivially check that these constraints are feasible and define the non-empty CSs $K(U_X)$, $K(U_Y)$, and $K(U_Z)$. This proves the M-compatibility of the endogenous PMF based on the data in Table \ref{tab:study}. CN inference eventually provides the PNS bounds. These bounds are computed in the twin network of Figure~\ref{fig:twinpearl} by first performing the interventions in the two copies of the $X$ variables and then jointly querying the $Y$ variables. The resulting bounds coincides with the one presented by \cite{exact2021causes} on the basis of their formulae. 
\end{example}

\begin{table}[htp!]
\centering
\begin{tabular}{cccr}
\toprule
Gender ($Z$) & Treatment ($X$)&Recovery ($Y$)&$\#$\\
\midrule
0&0&0&2\\
0&0&1&114\\
0&1&0&41\\
0&1&1&313\\
1&0&0&107\\
1&0&1&13\\
1&1&0&109\\
1&1&1&1\\
\bottomrule
\end{tabular}
\caption{Data from an observational study involving three Boolean variables \citep[Section~4.1]{exact2021causes}. The state equal to one means \emph{female} for $Z$, \emph{treated} for $X$ and \emph{recovered} for $Y$.}
\label{tab:study}
\end{table}

The M-compatibility with the endogenous PMF induced by the data in Table \ref{tab:study} for the Markovian model in Example~\ref{ex:pearl} reflects a general result about M-compatibility and Markovianity.

\begin{theorem}\label{th:conservative}
In a canonical Markovian PSCM M over $(\bm{U},\bm{V})$ any $P(\bm{V})$ is M-compatible.
\end{theorem}

In practice, by Theorem~\ref{th:conservative}, if the SEs of a Markovian SCM are not available and we adopt a canonical specification, compatibility is guaranteed.

The bounds in \cite{exact2021causes} correctly predict the right numbers we get in Example~\ref{ex:pearl}. But we are not only assuming that the data factorise according to the graph in Figure~\ref{fig:pearl}, as the mentioned paper does. We are also assuming that the specification is canonical (i.e., all the mechanisms are possible). What happens if we do not---considered that the latter is not an assumption required to derive the bounds. This is discussed in the next example, still based on the setup of Example~\ref{ex:pearl}.

\begin{example}\label{ex:pearl3}
Assume that, thanks to some expert knowledge, the state $U_X=3$ is deemed impossible, this meaning that a deterministic mechanism forcing the treatment of all the patients is considered unrealistic. Such an assumption preserves the separate surjectivity of $f_X$ and also the joint surjectivity of the model. In other words there is at least a joint state of $(U_X,U_Y,U_Z)$ inducing any possible joint state of $(X,Y,Z)$. Under this additional information, the negation of Equation~\eqref{eq:c1} rewrites as:
\begin{equation}\label{eq:c3}
P(U_X=0)=P(X=1|Z=0)=\frac{354}{470}
\,,
\end{equation}
and this is clearly inconsistent with Equation~\eqref{eq:c2}. 
\end{example}

This means that the model obtained by dropping state $U=3$ cannot generate the distribution that we see in the data: namely, the data are incompatible with such a non-canonical model. Note also that this shows that the endogenous distribution of the data can factorise according to the graph, while at the same time, it may be incompatible with the related model. This may happen because the latter form of incompatibility depends on the SEs, not only on the graph.

In fact, as shown by the next example, M-incompatibility is subtle in that it may be based on an endogenous probability that is inconsistent with the very model used to produce the data!

\begin{example}\label{ex:pearl4}
In the same setup of Example~\ref{ex:pearl} assume that some expert tells us that, out the sixteen states of $U_Y$, only three of them are possible, namely those indexing the following logical relations:
\begin{eqnarray}
f_Y(X,Z,U_Y=0)&=&X \vee \neg Z\,,\\
f_Y(X,Z,U_Y=1)&=&\neg X \vee \neg Z\,,\\
f_Y(X,Z,U_Y=2)&=&\neg X \wedge Z\,.
\end{eqnarray}
Note that this preserves surjectivity. Call $M'$ the reduced SCM that we obtain from the canonical specification (in M) by keeping only the values $0,1,2$ both for $U_X$ and $U_Y$. Consider the following distributions for the exogenous variables in $M'$: $P(U_Y):=[0.47, 0.439, 0.091]$, $P(U_X):=[0.677,0.000,0.323]$, and $P(U_Z)$ taking the same values of $P(Z)$ as in the case of M. Because of the previous discussion, the data in Table~\ref{tab:study} are incompatible with $M'$. And yet they can be produced by $M'$, because $P(X,Y,Z)$ is a positive distribution under $M'$. So any data can be generated---with different log-likelihoods. In particular, the ratio between the log-likelihood of $M'$ and that of M is $0.71$: thus it turns out that it is not at all unlikely to produce the data in Table~\ref{tab:study} with model $M'$. And since those data factorise according to the graph, as before, we should be allowed to apply the bounds as before, claiming that the upper bound of PNS for $M'$ is $0.015$. But the PNS for $M'$ is $0.15$, 10 times above the bound. And so the bounds of \citet[Section~4.1]{exact2021causes} fail here.
\end{example}

They fail because `factorising according to the graph' is not strong enough an assumption to derive bounds; we should ask for M-compatibility. Stated differently, whenever we produce formulae, or algorithms, for computing counterfactuals, we should make sure that M-compatibility holds, otherwise the results will be unwarranted. 


At this point, one might be tempted to escape the problem of testing M-compatibility in absence of SEs (i.e., in presence of data and the causal graph alone) by using a canonical specification of the SEs, given that the resulting model is always compatible with the data (cf. Theorem~\ref{th:conservative}). Even more so that recent efforts \citep{duarte2021, zhang2021} appear to provide a canonical specification for general PSCMs; this would seem to enable doing general counterfactual inference without SEs. Yet, canonical specifications do not seem to provide us with a safe way out, as shown next.



Given a canonical Markovian PSCM M, any other Markovian PSCM $M'$ over the same endogenous variables and graph can be defined by dropping some values of the exogenous variables, this being equivalent to remove from the SEs the corresponding mechanisms.
That is, $M'$ is defined via sets $\{\Omega_U'\}_{U\in\bm{U}}$ such that, for each $U\in\bm{U}$, $\Omega_U'\subseteq\Omega_U$ are the states of $U$ indexing the deterministic relations of the corresponding SE of $M'$. As a consequence we can regard a canonical PSCM as the set of all the possible non-canonical PSCMs. Some of these may be incompatible with the data, though. So how does inference in the canonical PSCM and in the set of non-canonical PSCMs relate to each other?   

To answer this question, let us say that $M$ \emph{embeds} $M'$ if and only there is at least a $P(\bm{U})\in\mathcal{K}$
(see Theorem~\ref{th:unimod} for the definition of $\mathcal{K}$) assigning $P(U=u)=0$ to all values $u$, of all exogenous variables $U$, corresponding to SEs that should be dropped to obtain $M'$ from $M$.\footnote{The analogous definition in \citet{zaffalon2021} mistakenly reported $P(U\in\Omega'_U)>0$ instead of $P(U\in\Omega'_U)=1$ as in this paper.} In other words, embedding a sub-model $M'$ means to `soft'-drop the SEs that do not belong to it via zero probabilities. The following holds:

\begin{theorem}\label{cor:discarding}
A Markovian canonical PSCM $M$ cannot embed incompatible models.
\end{theorem}

As a consequence, running an inference algorithm in a canonical model corresponds to \emph{automatically discarding} the incompatible models while using only the compatible ones. The important implication here is that if the true underlying model is not compatible with the available data, the results obtained by using the canonical model will be unwarranted as an approximation to the actual one. This is in fact the reason why in Example~\ref{ex:pearl4} the PNS interval obtained by the canonical model does not contain the actual value of PNS.

Overall, the lesson appears to be that we cannot have guaranteed bounds without knowing the SEs of the underlying SCM. The relation of M-compatibility to sample size appears therefore to be a subject that needs to be studied more deeply in order to at least provide probabilistic guarantees on the delivered bounds.
\section{Causal Expectation Maximisation}\label{sec:emcc}
In spite of the hardness of causal inference (cf. Theorem~\ref{th:hard}), the mapping to CN inference derived in the previous section may allow to take advantage of existing CN algorithms. The limitations are those related to non-quasi-Markovian models and to the intrinsic hardness of CN inference. Nevertheless, we show in this section that it is possible to leverage the particular optimisation required by causal queries, which leads to an iterative EM approach that provides an approximate range for partially identifiable queries for general PSCMs (Section \ref{sec:em}). We also derive credible intervals to characterise the quality of such an approximation (Section \ref{sec:credible}).

\subsection{EM for Causal Computations}\label{sec:em}
Consider a partially identifiable query as in Equation \eqref{eq:bounds}. 
We already noticed that the set of FSCMs $\mathcal{M}_{M,P}$ is in a one-to-one correspondence with the elements of the set $\mathcal{K}$ considered by Theorem~\ref{th:unimod}.
Thus, in principle, if $\mathcal{K}$ is available, one could compute the bounds of a query by optimising the corresponding function of $P(\bm{U})$ over $\mathcal{K}$. 

In practice coping with partial identifiability is demanding even with simple topologies and queries (cf. Theorem~\ref{th:hard}) and we should generally consider methods to compute approximate bounds. As a matter of terminology, here and in the following we shall call \emph{range} the values spanned by the counterfactual query inside the approximate bounds. As mentioned already in the Introduction, the range is an inner approximation of the actual counterfactual interval as identified by the exact bounds.

Corollary \ref{cor:k} provides a good match w.r.t. developing approximations: it tells us that sampling the global optimum points of the log-likelihood corresponds to sampling from, and hence approximating, $\mathcal{K}$. And the crucial observation is that we can easily sample the optimum points of the log-likelihood with the \emph{expectation-maximisation} (EM) scheme \citep{dempster1977maximum}: in fact, exogenous variables are missing at random in $\mathcal{D}$, just because they are latent (missing with probability one).

In particular, given an initialisation $P_0(\bm{U})$, the EM algorithm consists in regarding the posterior probability $P_0(u|\bm{v})$ as a \emph{pseudo-count} for $(u,\bm{v})$, for each $\bm{v}\in\mathcal{D}$,  $u\in\Omega_U$ and $U\in\bm{U}$ (E-step). A new estimate is consequently obtained as $P_1(u):=|\mathcal{D}|^{-1}\sum_{\bm{v}\in\mathcal{D}} P_0(u|\bm{v})$ (M-step). This scheme, called EMCC (\emph{EM for Causal Computation}) is iterated until convergence. 
Algorithm~\ref{alg:cem} depicts the EMCC pseudo-code.

Subroutine ${\tt initialise}$ (line 1) provides a random initialisation of the exogenous PMFs, while
${\tt components}$ (line 2) finds the c-components of M. A restriction of the dataset to the variables in the $c$-th component is achieved in line~4. In line~8 d-separation properties allow to replace $P(U|\bm{v})$ with $P(U|\bm{w}^{c})$ when $U\in\bm{U}^{c}$. Since $P_{t+1}$ gets a higher log-likelihood than $P_t$ \cite[Theorem~19.3]{koller2009}, we adopt likelihood stationarity as a stopping criterion to decide convergence (line~10). As a side remark, note that each iteration of the loop in line~3 of Algorithm~\ref{alg:cem} can be executed in parallel to the others because of the d-separation among c-components. Something similar can be done at the dataset level when computing, by standard BN algorithms, the queries in line~8. 

\begin{algorithm}[htp!]
{\color{myblueR2}\caption{EM for Causal Computation (EMCC).}\label{alg:cem}
\SetKwInOut{Input}{input}\SetKwInOut{Output}{output}
\Input{A PSCM M over $(\bm{U},\bm{V})$ and a dataset $\mathcal{D}$ of observations of $\bm{V}$}
\Output{$P(\mathbf{U})$}
$P_0(\mathbf{U}) \leftarrow {\tt initialise}(M)$\;
$\{\bm{U}^{c}, \bm{V}^{c}\}_{c\in\mathcal{C}} \leftarrow {\tt components}(M)$\;
\For{$c \in \mathcal{C}$}{
$\mathcal{D}^{c} \leftarrow \mathcal{D}^{\downarrow \bm{W}^{c}}$\;
$t \leftarrow 0$\;
\Repeat{$l(P_{t+1}(\mathbf{U}^c))= l(P_{t}(\mathbf{U}^c))$}{
\For{$U\in \bm{U}^{c}$}{
$P_{t+1}(U) \leftarrow |\mathcal{D}|^{-1}\sum_{\bm{w}^{c} \in \mathcal{D}^{c}} P_t(U|\bm{w}^{c})$
$t \leftarrow t+1$}}}}
\end{algorithm}

Multiple EMCC runs, started with different seeds, will yield an approximating subset of $\mathcal{K}$. 
It is known \citep{Wu_1983} that the EM algorithm only converges to stationary points of the likelihood. In practice, after the convergence of EM we evaluate the log-likelihood and, if a value smaller than the global maximum is achieved, we reject the point. Empirically we have never observed a case of convergence to a stationary point that is either a global maximum or a saddle point.

\subsection{Deriving Credible Intervals for EMCC Inferences}\label{sec:credible}
In this section we characterise the accuracy of our EMCC procedure in terms of credible intervals. 
Let us consider a, possibly partially identifiable, query, whose exact bounds defined as in Equation \eqref{eq:bounds} are $[a^*,b^*]$. In this section we assume that $[a^*,b^*] \subseteq [0,1]$, i.e., the query results in a probability value, to simplify the notation. The results listed below, however can be easily extended to the general case of any bounded query $[a^*,b^*]$.

Say that $k$ EMCC runs have been executed to approximate the query and denote the result as $\rho:=\{\pi_i\}_{i=1}^k$. Let $a:=\min_{i=1}^k \pi_i$ and $b:=\max_{i=1}^k \pi_i$ denote their approximations and let $L:=b-a$ be the size of the range. By construction, we have $a^*\leq a \leq b \leq b^*$. To evaluate the quality of such an inner approximation we compute the probability of covering $[a^*,b^*]$ if we assume that both $a$ and $b$ are subject to an error of $\epsilon$ relative to $L$. The probability depends on how the EMCC outputs are distributed:

\begin{theorem}\label{th:newbounds}
Assume that the EMCC runs in $\rho$ are distributed as a four parameter beta distribution, i.e., $\pi_i \sim Beta(\alpha,\beta,a^*, b^*)$, for each $i=1, \ldots, k$. 
The following equality holds:
\begin{equation}\label{eq:confidenceBeta}
P\left(
a-\varepsilon L\leq
a^*\leq b^* \leq b+\varepsilon L\,\bigg|\, \rho \right) 
=  
\dfrac{\displaystyle\int_{0}^{\delta_b/2}\int_{0}^{\delta_a/2} P(x,y; L,\alpha, \beta,k) \, \mathrm{d}x \, \mathrm{d}y}{\displaystyle\int_{0}^{a+(1-b)}\int_{0}^{a+(1-b)-y}P(x,y; L,\alpha, \beta,k) \, \mathrm{d}x \, \mathrm{d}y}\,,
\end{equation}
where $P(x,y; L,\alpha, \beta, k)$ is equal to
\begin{equation}\label{eq:newbounds}
\left(\dfrac{(L+x)^\alpha \twoFone(\alpha,1-\beta,\alpha+1,\frac{L+x}{L+x+y})- x^\alpha \twoFone(\alpha,1-\beta,\alpha+1,\frac{x}{L+x+y})}{\alpha (L+x+y)^{\alpha} B(\alpha,\beta)}\right)^k\,,
\end{equation}
$\twoFone$ is the Gaussian (ordinary) hypergeometric function, $B(\alpha,\beta)$ is the beta function evaluated at $\alpha,\beta>0$ and:
\begin{eqnarray}
\label{eq:delta_a}
\delta_a &:=&\left\{\begin{array}{ll} 2L\varepsilon &\mathrm{if}\,\, \varepsilon \leq \frac{a}{L}\\ 2a&\mathrm{otherwise}\end{array}\right.\\
\delta_b &:=&\left\{\begin{array}{ll} 2L\varepsilon&\mathrm{if}\, \varepsilon \leq \frac{1-b}{L}\\ 2(1-b)&\mathrm{otherwise}.\end{array}\right.
\label{eq:delta_b}
\end{eqnarray}
Note that Equation~\eqref{eq:confidenceBeta} is a valid probability if $\varepsilon$ is such that:
$\delta_a \leq 2(1-L) - \delta_b$ 
and
$\delta_b \leq 1-L$. 
\end{theorem}

There are some notable special cases to the above procedure for which a simplified version of Equation~\eqref{eq:confidenceBeta} can be derived.

\begin{corollary}\label{cor:boundaequalb}
For $a=0$, we have $a^*=0$, $L=b$, and $\pi_i \sim Beta(\alpha,\beta,0, b^*)$, for each $i=1,\ldots,k$, and hence:
\begin{equation}\label{eq:confidenceBeta2}
P\left(
 b^* \leq b+\varepsilon L\,\bigg|\, \rho \right) 
=  
\dfrac{\displaystyle\int_{0}^{\delta/2} P(y; L,\alpha, \beta,k)  \, \mathrm{d}y}{\displaystyle\int_{0}^{(1-b)}P(y; L,\alpha, \beta,k) \, \mathrm{d}y}\,,
\end{equation}
with $\delta =2L\varepsilon$ for $\varepsilon \leq \frac{1-b}{L}$, where $P(y; L,\alpha, \beta, k)$ is equal to
\begin{equation}\label{eq:newbounds3}
\left(\dfrac{L^\alpha \twoFone(\alpha,1-\beta,\alpha+1,\frac{L}{L+y})}{\alpha (L+y)^{\alpha} B(\alpha,\beta)}\right)^k\,.
\end{equation}
Similarly, for $b=1$, we have $b^*=1$, $L=1-a$, and $\pi_i \sim Beta(\alpha,\beta,a^*,1)$, for each $i=1, \ldots,k$ and hence:
\begin{equation}\label{eq:confidenceBeta3}
P\left(
a^* \geq a-\varepsilon L\,\bigg|\, \rho \right) 
=  
\dfrac{\displaystyle\int_{0}^{\delta/2} P(x; L,\alpha, \beta,k)  \, \mathrm{d}x}{\displaystyle\int_{0}^{a}P(x; L,\alpha, \beta,k) \, \mathrm{d}x}\,,
\end{equation}
with $\delta =2L\varepsilon$ for $\varepsilon \leq \frac{a}{L}$, where $P(x; L,\alpha, \beta, k)$ is equal to
\begin{equation}\label{eq:newbounds2}
\left(\dfrac{(L+x)^\alpha \twoFone(\alpha,1-\beta,\alpha+1,\frac{L+x}{1+L+x})- x^\alpha \twoFone(\alpha,1-\beta,\alpha+1,\frac{x}{1+L+x})}{\alpha (1+L+x)^{\alpha} B(\alpha,\beta)}\right)^k\,.
\end{equation}

Finally, for $a=b$, i.e., all $k$ runs in $\rho$ are equal, then:
\begin{equation}\label{eq:identifiable}
P(a^*=b^*|\rho) =1+ 9/3^k-8/2^k\,.    
\end{equation}
\end{corollary}
The last case in the corollary above implies that nine equal runs make identifiability probable with 99\% confidence.

Theorem~\ref{th:newbounds} is a proper extension of the original EMCC characterisation \citep{zaffalon2021}, as proved by the following result.
\begin{corollary}\label{cor:newbounds}
If the EMCC runs in $\rho$ are uniformly distributed in $[a^*, b^*]$,
i.e., for each $i=1, \ldots, k$, $\pi_i \sim Beta(1,1,a^*, b^*)$, then we get:
\begin{equation}\label{eq:bounds_uniform}
P\left(
a-\varepsilon L\leq
a^*\leq b^* \leq b+\varepsilon L\,\bigg|\, \rho \right) =
\frac{
1+(1+2\varepsilon)^{2-k}-2(1+\varepsilon)^{2-k}}{
(1-L^{k-2})
-(k-2)(1-L)L^{k-2}}
\,.
\end{equation}
\end{corollary}

In practice we can use the theorems above as stopping criterion for the EMCC procedure. 
After any $k$ EMCC runs, we compute $a$ and $b$ and, given a relative error $\epsilon$ we regard as acceptable, compute the probability in Equation~\eqref{eq:confidenceBeta}. If the corresponding probability is sufficiently high, we stop iterating EMCC, otherwise we keep iterating the procedure to collect new points for $\rho$ and achieve greater probabilities. 

Note that the computation of the probability in Equation~\eqref{eq:confidenceBeta} requires estimating the parameters $\alpha$ and $\beta$ of the distribution $Beta(\alpha,\beta, a^*, b^*)$. We estimate them via a maximum likelihood procedure over the $k$ values collected in $\rho$. Moreover, since $a^*,b^*$ are unknown, we also need to choose the range of the $Beta$ distribution. Here we set $Beta(\alpha,\beta, a-\varepsilon L,  b+\varepsilon L)$.  The integrals in Equations~\eqref{eq:confidenceBeta}, \eqref{eq:confidenceBeta2} and \eqref{eq:confidenceBeta3} cannot be computed analytically and are thus estimated numerically.

\section{Empirical Validation}\label{sec:experiments}
We present here a numerical validation of our techniques to estimate the bounds of partially identifiable queries. This is achieved through simulations on an extensive benchmark of synthetic models (Section \ref{sec:exp}), a case study on palliative care (Section \ref{sec:triangolo}), and the analysis of a classical BN for medical diagnosis (Section \ref{sec:real}).

\subsection{Tests on Synthetic Models}\label{sec:exp}
Let us test both the CN mapping described in Section~\ref{sec:cn} and the EMCC algorithm introduced in Section~\ref{sec:emcc}. To do that we consider the computation of counterfactual, partially identifiable, queries on a benchmark of synthetic PSCMs.

\paragraph{Sampling FSCMs} For each experiment we first sample a `ground-truth' FSCM $M^*$. The causal graph of $M^*$ is sampled with the Erd\"os-R\'enyi model specialised to the case of directed acyclic graphs \citep{ide2004generating}. We set the maximum in-degree equal to three and the maximum out-degree equal to two, with the total number of nodes specified as an input. We cope with semi-Markovian models, this implying that only the root nodes are associated with the exogenous variables. Thus, if the sampled graph is such that there are non-root nodes without root parents, we add to the graph a new root node as a parent. Let $\mathcal{G}$ denote the causal graph obtained in this way. The root nodes of $\mathcal{G}$ correspond to the exogenous variables $\bm{U}$ of $M^*$, while the non-root nodes refer to the endogenous variables in $\bm{V}$. Just for the sake of simplicity, the variables in $\bm{V}$ are assumed Boolean. Regarding the cardinality of the exogenous variables and the SEs, for the c-components including a single exogenous variable we adopt a canonical specification. If two or more exogenous variables are in the same c-component, we set instead $|\Omega_U|=16$, for each $U$ in the component. {\color{myblue}{The SE of each endogenous variable in the c-component is obtained by sampling a surjective map determining the value of the endogenous from that of its endogenous parents, for each joint state of the exogenous parents.}} Let $M$ denote the PSCM obtained in this way. FSCM $M^*$ is eventually obtained by adding to $M$ a (uniformly sampled) random PMF $P^*(U)$ such that $P(U=u)>0$ for each $u\in\Omega_U$, for each $U \in \bm{U}$. Overall, we create a benchmark of $101$ FSCMs, $39$ of them being quasi-Markovian. The number of variables over the different models in the benchmark ranges from $5$ to $19$ (average $9.9$), while the exogenous cardinality ranges from $3$ to $256$ (average $30.2$) and the treewidth of $\mathcal{G}$ from $2$ to $4$ (average $2.9$).


\paragraph{M-compatibility} 
A dataset $\mathcal{D}$ of endogenous observations is sampled from $M^*$. The PSCM $M$ underlying FSCM $M^*$ and the dataset $\mathcal{D}$ (together with the endogenous BN quantified from $\mathcal{D}$) are the inputs of each experiment. To check the M-compatibility of the (joint PMF associated with the) endogenous BN, we first use the BN to compute $\lambda^*$ (see Equation~\eqref{eq:likdec}). After that, we perform a single EMCC run with $M$ and $\mathcal{D}$ as inputs to check whether, after convergence, the maximum value $\lambda^*$ is achieved by the likelihood (cf. Corollary~\ref{cor:k}). If this is not case, we sample more observations until M-compatibility is guaranteed. 

\paragraph{Queries and Ground Truth} As a partially identifiable (counterfactual) query, we consider a PNS having as \emph{cause} and \emph{effect} respectively the first and the last of the variables in $\bm{V}$, assuming them sorted in a topological order. The bounds $[a^*,b^*]$ can be computed only for quasi-Markovian models by exact CN inference. In practice this is feasible only with small models ($<10$ nodes). For the other models in the benchmark we use the EMCC (Algorithm~\ref{alg:cem}) with a high ($>300$) number $r$ of runs. Let $[a_r,b_r]$ denote the range induced by those $r$ runs. Taking a threshold probability $P^*:=0.99$, we check the minimum value $\epsilon^*$ of $\epsilon$ such that the probability in Equation~\eqref{eq:confidenceBeta2} is greater than or equal to $P^*$. This allows to further approximate the true bounds as $a^* \simeq a_r-\epsilon^* L$ and $b^* \simeq b_r+\epsilon^* L$. We take these as our `ground-truth' bounds to be compared with the inner approximation obtained by the EMCC with a smaller number of runs. For quasi-Markovian models, we also use the CN mapping (Algorithm \ref{alg:quasi}) linked with the approximate \emph{ApproxLP} algorithm for CN inference. Note that also this procedure provides inner approximations \citep{approxlp}. The quality of the (inner) approximation of $[a^*,b^*]$ provided by the range $[a,b]$ in output of one of our algorithms is described by the following relative root mean square error:
\begin{equation}\label{eq:rrmse}
\mathrm{RRMSE} := \sqrt{\frac{(a-a^*)^2+(b-b^*)^2}{2(b^*-a^*)^2}}\,,
\end{equation}
which is computed for each experiment. 

\paragraph{Implementation and Results} 
Both the ECMM and the CN mapping have been implemented within the CREDICI library for causal inference \citep{credici}.\footnote{\href{https://www.github.com/idsia/credici}{github.com/idsia/credici}.} CN inferences are computed instead by the CREMA library \citep{huber2020crema}, which can be directly imported in CREDICI. Note that the CREDICI library also includes scripts for model and data generation, as well as the computation of the integrals to obtain the credible intervals discussed in Section~\ref{sec:credible}. The scripts to reproduce the results discussed here and in Section \ref{sec:real} are stored in a separate repository, which also includes the case studies discussed in the next sections.\footnote{\href{https://www.github.com/IDSIA-papers/2023-IJAR-efficient-bounding}{github.com/IDSIA-papers/2023-IJAR-efficient-bounding}.}
The simulations are sequentially executed on an AMD EPYC-7542 32-Core Processor with 256 GB.

\begin{figure}[htp!]
\centering
\begin{tikzpicture}
\begin{axis}[width=0.7\textwidth,height=0.5\textwidth,name=border,ylabel={RRMSE},xmin=-1,xmax=19,xlabel={EMCC runs},
boxplot/draw direction=y,cycle list={{black}},xtick={0,2,4,6,8,10,12,14,16,18},xticklabels={20,40,60,80,100,120,140,160,180,200}]
\addplot+[boxplot prepared={draw position=0,lower whisker=0,lower quartile =0.039,median=0.125,upper quartile =0.198,upper whisker=0.384},] coordinates {};
\addplot+[boxplot prepared={draw position=2,lower whisker=0,lower quartile =0.02,median=0.097,upper quartile =0.167,upper whisker=0.317},] coordinates {};
\addplot+[boxplot prepared={draw position=4,lower whisker=0,lower quartile =0.017,median=0.066,upper quartile =0.143,upper whisker=0.258},] coordinates {};
\addplot+[boxplot prepared={draw position=6,lower whisker=0,lower quartile =0.015,median=0.057,upper quartile =0.14,upper whisker=0.257},] coordinates {};
\addplot+[boxplot prepared={draw position=8,lower whisker=0,lower quartile =0.015,median=0.057,upper quartile =0.131,upper whisker=0.253},] coordinates {};
\addplot+[boxplot prepared={draw position=10,lower whisker=0,lower quartile =0.012,median=0.056,upper quartile =0.117,upper whisker=0.253},] coordinates {};
\addplot+[boxplot prepared={draw position=12,lower whisker=0,lower quartile =0.01,median=0.056,upper quartile =0.114,upper whisker=0.253},] coordinates {};
\addplot+[boxplot prepared={draw position=14,lower whisker=0,lower quartile =0.01,median=0.056,upper quartile =0.114,upper whisker=0.253},] coordinates {};
\addplot+[boxplot prepared={draw position=16,lower whisker=0,lower quartile =0.006,median=0.051,upper quartile =0.108,upper whisker=0.253},] coordinates {};
\addplot+[boxplot prepared={draw position=18,lower whisker=0,lower quartile =0.006,median=0.043,upper quartile =0.105,upper whisker= 0.223},] coordinates {};
\addplot[mark=o] coordinates {(18,0.253)};
\end{axis}
\end{tikzpicture}
\caption{RRMSE vs. EMCC runs.}\label{fig:emcc}
\end{figure}
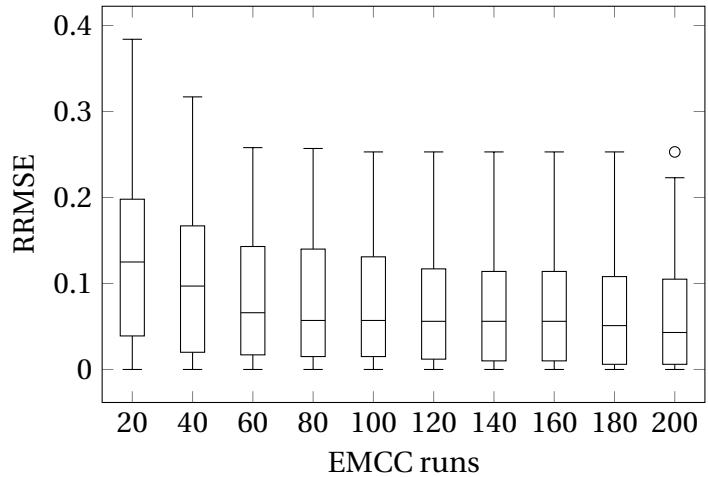

The boxplots in Figure~\ref{fig:emcc} describe the RRMSE of the ranges obtained by the EMCC for an increasing number of runs. As expected, the accuracy of our ranges increases with the number of EMCC runs. With less than $200$ runs, we already obtain an RRMSE$<0.1$ in most of the cases. With many runs and small errors, it becomes more difficult for the EMCC to further expand its ranges: this might explain the slower improvements in the right part of Figure~\ref{fig:emcc}.

\begin{figure}[htp!]
\centering
\begin{tikzpicture}
\begin{axis}[width=0.6\textwidth,height=0.4\textwidth,name=border,ylabel={RRMSE},
boxplot/draw direction=y,cycle list={{black}},xtick={0,2},xticklabels={EMCC,CN (ApproxLP)}]
\addplot+[boxplot prepared={draw position=0,lower whisker=0.004452026292919134,lower quartile =0.07642522614224659,median=0.10457551382971574,upper quartile =0.16894896545695987,upper whisker=0.2529071993934107},] coordinates {};
\addplot+[boxplot prepared={draw position=2,lower whisker=0.00024074641722872088,lower quartile =0.029770309016960587,median=0.05126347053781914,upper quartile =0.08433357925769026,upper whisker=0.13651768227571356},] coordinates {};
\addplot[mark=o] coordinates {(2,0.19505408)};
\addplot[mark=o] coordinates {(2,0.18175391)};
\end{axis}
\end{tikzpicture}
\caption{EMCC ($r=200$) vs. approximate CN approach on quasi-Markovian models.}\label{fig:emccvscn}
\end{figure}
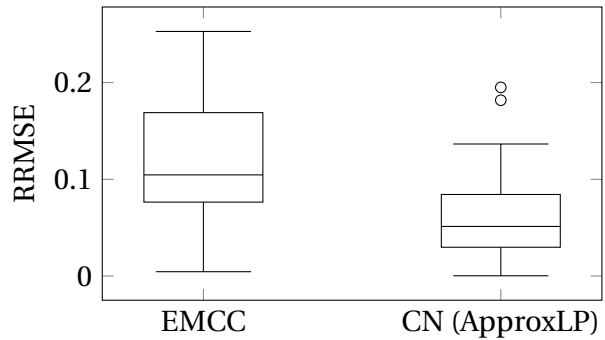

Figure~\ref{fig:emccvscn} depicts instead a comparison between EMCC and CN approaches. The ApproxLP algorithm invoked by the CN method is slightly more accurate, but, contrary to the EMCC, its direct application is restricted to quasi-Markovian models. Thus, we might reasonably regard the EMCC as the algorithm of choice to compute the ranges of partially identifiable queries.

\subsection{A Counterfactual Analysis in Palliative Care}\label{sec:triangolo}
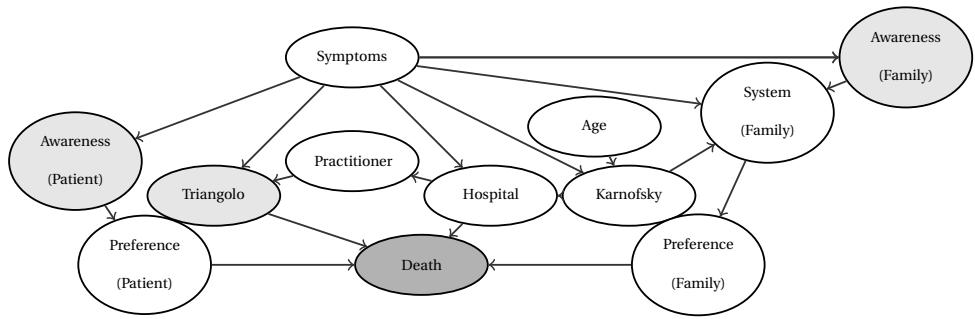
\begin{figure}[htp!]
\centering
\begin{tikzpicture}[scale=0.92]
\node[nodo] (s)  at (0,1.0) {\tiny Symptoms};
\node[nodo] (a)  at (3.5,0.0) {\tiny Age};
\node[nodo2] (ap)  at (-4.0,-0.5) {\tiny Awareness (Patient)};
\node[nodo2] (t)  at (-2,-1.0) {\tiny Triangolo};
\node[nodo] (p)  at (0,-0.5) {\tiny Practitioner};
\node[nodo] (h)  at (+2,-1.0) {\tiny Hospital};
\node[nodo] (k)  at (+4.0,-1.0) {\tiny Karnofsky};
\node[nodo] (sf)  at (+6,0.2) {\tiny System (Family)};
\node[nodo2] (af)  at (+8.0,1.0) {\tiny Awareness (Family)};
\node[nodo] (pf)  at (5.0,-2.0) {\tiny Preference (Family)};
\node[nodo] (pp)  at (-3.0,-2.0) {\tiny Preference (Patient)};
\node[nodo3] (d)  at (1.0,-2.0) {\tiny Death};
\draw[arco] (s) -- (ap);
\draw[arco] (s) -- (af);
\draw[arco] (s) -- (t);
\draw[arco] (s) -- (h);
\draw[arco] (h) -- (p);
\draw[arco] (p) -- (t);
\draw[arco] (k) -- (h);
\draw[arco] (k) -- (sf);
\draw[arco] (af) -- (sf);
\draw[arco] (s) -- (k);
\draw[arco] (s) -- (sf);
\draw[arco] (s) -- (af);
\draw[arco] (sf) -- (pf);
\draw[arco] (ap) -- (pp);
\draw[arco] (pp) -- (d);
\draw[arco] (pf) -- (d);
\draw[arco] (a) -- (k);
\draw[arco] (t) -- (d);
\draw[arco] (h) -- (d);
\end{tikzpicture}
\caption{The model to study preferences about the place of death in oncology patients.}
\label{fig:triangolo}
\end{figure}

Figure~\ref{fig:triangolo} represents a causal model for the study of terminally ill cancer patients' preferences with respect to their place of death: home or hospital. In fact most patients prefer to die at home, but the majority actually die in institutional settings. The study aimed at understanding interventions by health care professionals that can facilitate dying at home. The graph corresponds to the network proposed by \cite{kern2020impact} reduced to the subset of variables for which data were available---variables have been binarised too. 

One can intervene on three variables in the network (light grey nodes in Figure~\ref{fig:triangolo}): the patient's and the family's \emph{awareness} of death (which involves communication with the doctors); and home assistance (provided by the \emph{Triangolo} association). Our goal is to use the tools of causal analysis to understand what is the most important variable on which to act, and we measure importance by the PNS values having the variable \emph{death} as effect variable (dark grey node in Figure~\ref{fig:triangolo}). The idea is that the one with highest PNS will lead to the highest increase of people that can die at home by its very reason.

To this end, we turn the casual graph into a Markovian SCM by adding an exogenous variable to each node. The model is taken Markovian as a consequence of the fact that all the potential confounders have been explicitly represented in the causal graph. As for SEs,  we stick to the canonical representation since we want to be least-committal w.r.t. the true underlying mechanisms. Note that this induces a high cardinality for the exogenous variable associated with node \emph{death} as, by Equation~\eqref{eq:canonical}, we have $|\Omega_{U_\mathrm{death}}|=2^{16}$. We generate a sample of endogenous data from the original network model (we have no access to the patients' original data). The high cardinality of $\Omega_{U_{\mathrm{death}}}$ prevents an application of the CN algorithms, while the EMCC can be executed and we obtain our PNSs based on $160$ EMCC runs (EM convergence achieved in around 500 iterations on average) in approximately 75 minutes. More specifically, $[0.30, 0.31]$ is the resulting range when \emph{Triangolo} is the intervened variables; similarly, \emph{patient's awareness} gives $[0.03,0.10]$ and \emph{family's awareness} $[0.06, 0.10]$. The conclusion is that one should clearly act on Triangolo first: for instance, by making Triangolo available to all patients, we should expect a reduction of people at the hospital by $30$\%. This would save money too, and would allow politicians to do economic considerations as to which amount it is even economically profitable to fund Triangolo, and have patients die at home, rather than spending more to have patients die at the hospital.

\subsection{A Counterfactual Analysis with the Asia Network}\label{sec:real}
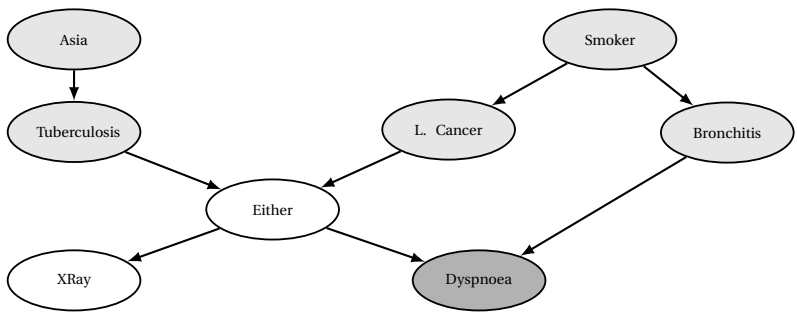
\begin{figure}[htp]
\centering
\begin{tikzpicture}
\tikzstyle{connect}=[-latex, thick]
\newcommand{\xfactor}{*0.015};
\newcommand{\yfactor}{*-0.008};
\node[nodo2] (Asia) at (300\xfactor, 150\yfactor)  {\tiny Asia};
\node[nodo2] (Smoker) at (775\xfactor, 150\yfactor)  {\tiny Smoker};
\node[nodo2] (Tuberculosis) at (300\xfactor, 304\yfactor)  {\tiny Tuberculosis};
\node[nodo2] (LungCancer) at (632\xfactor, 300\yfactor)  {\tiny L. Cancer};
\node[nodo2] (Bronchitis) at (879\xfactor, 305\yfactor)  {\tiny Bronchitis};
\node[nodo] (Either) at (476\xfactor, 433\yfactor)  {\tiny Either};
\node[nodo3] (Dyspnoea) at (659\xfactor, 551\yfactor)  {\tiny Dyspnoea};
\node[nodo] (XRay) at (300\xfactor, 551\yfactor)  {\tiny XRay};
\path (Asia) [connect]  edge (Tuberculosis);
\path (Bronchitis) [connect]  edge (Dyspnoea);
\path (Either) [connect]  edge (Dyspnoea);
\path (Either) [connect]  edge (XRay);
\path (LungCancer) [connect]  edge (Either);
\path (Smoker) [connect]  edge (Bronchitis);
\path (Smoker) [connect]  edge (LungCancer);
\path (Tuberculosis) [connect]  edge (Either);
\end{tikzpicture}	
\caption{The graph of the Asia Bayesian network used for the causal analysis.}
\label{fig:asia}
\end{figure}

In line with the preceding section, we perform a causal analysis based on a classical BN model, which is also related to a medical domain. We consider the \emph{Asia} BN, whose graph is depicted in Figure~\ref{fig:asia}. In this case, the aim is to determine the foremost causes to the occurrence of \emph{Dyspnoea}. For this purpose, we compute the PNS values having the variable \emph{Dyspnoea} as effect variable, and the variables \emph{Bronchitis}, \emph{Lung Cancer}, \emph{Tuberculosis}, \emph{Smoker} and \emph{Asia} as causes.

As in the study on palliative care, we obtain from the causal graph a Markovian SCM by introducing an exogenous variable to each node and considering canonical SEs. A dataset of $5,000$ instances is sampled from the initial BN. PNS ranges are obtained by means of the CN mapping (Algorithm \ref{alg:simple}) linked to exact and approximate (ApproxLP) CN algorithms. EMCC ranges are also computed with $r=71$ runs. The PNS ranges rounded to the second decimal place are in Table \ref{tab:asia}. Both approximate methods are very accurate. \emph{Bronchitis} appears as the most relevant variable for an intervention aiming to avoid \emph{Dyspnoea}, while \emph{Asia} is the least relevant.

\begin{table}[h!bt]
\centering
\begin{tabular}{lccc}
\hline
Cause& CN (Exact) & CN (Approx) & EMCC \\
\hline
Bronchitis& $[0.65,0.76]$&$[0.65,0.76]$&$[0.65,0.76]$\\ 
Lung Cancer	& $[0.37,0.48]$&$[0.37,0.48]$& $[0.37,0.48]$\\ 
Tuberculosis& $[0.36,0.46]$&$[0.36,0.46]$&$[0.36,0.46]$\\ 
Smoker& $[0.23,0.53]$&$[0.24,0.52]$&$[0.24,0.51]$\\ 
Asia&$[0.01,0.02]$&$[0.01,0.02]$&$[0.01,0.02]$\\ 
\hline
\end{tabular}
\caption{PNS ranges about the effect on \emph{Dyspnoea} of different causes for the model in Figure \ref{fig:asia}.}\label{tab:asia}
\end{table}

\section{Conclusions}\label{sec:conc}
We have presented two algorithms to do counterfactual inference in partially specified structural causal models: the first is based on a mapping to credal networks while the second on an EM scheme. The CN one is limited to quasi-Markovian models; it delivers exact inference for relatively small models and approximate inference otherwise. The EMCC algorithm is just approximate but works on any semi-Markovian SCM. Both algorithms appear to work rather efficiently and accurately. We have empirically verified this via a structured experimental validation based also on credible intervals to establish the quality of solutions obtained. In spite of a few alternative approaches in the literature to solve this kind of problems, we seem to be the first to do a systematic evaluation of the proposed algorithms as well as the first to publicly release the related code. We regard this as a relevant contribution of this work in order to start establishing fair comparisons of diverse approaches; in particular the NP-hardness of the involved problems, proved here, implies that solutions will have to be pursued via heuristics in general, which should be compared on standard benchmarks.
 
 Both algorithms assume that endogenous nodes are categorical, and both assume that structural equations are given---along with a dataset for the endogenous variables. Let us stress that the requirement that SEs are given is not as stringent as it might seem, given that we can produce the needed SEs by a preprocessing step if the actual ones are not available: this is possible thanks to recent work \citep{duarte2021,zhang2021} that has introduced a `canonical' specification of the SEs. This can be understood as a least-committal specification that, loosely speaking, can be used without loss of generality; the implication is that the output intervals will tend to be weaker compared to the case where the actual SEs are given. In this sense, our work is therefore as general as the works that do not assume the SEs to be given.

Another contribution of the present work is a theoretical analysis about the potential logical incompatibility of an SCM with (the empirical distribution inferred from) a dataset; we have called this M-incompatibility. We have observed that there are cases where M-incompatibility prevents one from obtaining guaranteed counterfactual bounds in absence of knowledge about the SEs, even when one uses the canonical specification. It is a problem that in principle can affect any method that tries to deliver bounds without knowing the actual SEs. It is unclear to us how severe this problem is in practice. For the time being we are pointing to a problem that does not seem to have obtained the due attention in the literature. We leave to future research a dedicated analysis that clarifies the problem in detail.

{\color{myblue}{We regard the EMCC in particular as a very practical algorithm to compute counterfactual inference, which---in this respect---does not seem to have many competitors at the moment. There are however many directions to improve it starting from the work in this paper. Some recent papers have already explored a few of these: \cite{zaffalon2023b} have extended the EMCC to  handle multiple data sources, of experimental and observational data, along with possible selection biases; \cite{huber2023a} have instead sped the EMCC up by one order of magnitude compared to the implementation described here, by a circuital compilation and some parallelisation steps. Note that the EMCC is query-agnostic at the moment: it aims at quantifying the uncertainty of the exogenous variables without targeting a specific counterfactual query. This means that once the uncertainty is quantified, any counterfactual query can be approximately solved without having to reconsider the data. However in some situations it may be more efficient to use query-specific algorithms that evaluate the uncertainty of the exogenous variables only in relation to a given query. This would entail using some local optimisation that would more efficiently find the counterfactual range. Finally, another useful direction would be to extend the EMCC to continuous domains.}}
\appendix

\section{Proofs}\label{app:proofs}
In this technical appendix we gather the proofs of the theorems presented in the paper together with some necessary lemma.

\begin{lemma}\label{lem:tian}
Equation~\eqref{eq:ident} is equivalent to have the following equation:
\begin{equation}\label{eq:tian}
\sum_{\bm{u}^c\in\Omega_{\bm{U}}^c} \left[ \prod_{U\in\bm{U}^c} P(u) \cdot \prod_{V\in\bm{V}^c} P(v|\mathrm{pa}(V)) \right] = \prod_{V\in\bm{V}^c} P(v|\bm{w}_V)\,,
\end{equation}
satisfied for each $c \in \mathcal{C}$.
\begin{proof}
The result has been proved by \cite{tian2002studies}. Equation~\eqref{eq:tian} corresponds to Equation (4.41) in that work.
\end{proof}
\end{lemma}

\paragraph{\bf Proof of Theorem \ref{th:unimod}}
Let us first consider the case $\mathcal{K} \neq \emptyset$. Take $\theta_{\bm{U}}\in\mathcal{K}$. We have Equation~\eqref{eq:tian} by Lemma~\ref{lem:tian}. Putting this in Equation~\eqref{eq:liku} we get:
\begin{equation}\label{eq:likdec2}
l(\theta_{\bm{U}}) = \sum_{c\in\mathcal{C}} \sum_{\bm{w}^c} n(\bm{w}^c) 
\sum_{V\in\bm{V}^c}
\log P(v|\bm{w}_V)\,.
\end{equation}
Equation~\eqref{eq:likdec2} corresponds to Equation~\eqref{eq:likdec} with the probabilities in Equation~\eqref{eq:freqs} that are giving the global maximum $\lambda^*$. This proves the sufficient condition. To prove the necessary condition, consider again Equation~\eqref{eq:tian}, which is equivalent to Equation~\eqref{eq:ident}. Assume, \emph{ad absurdum}, that there is a $\theta_{\bm{U}}\not\in\mathcal{K}$ such that the log-likelihood in Equation~\eqref{eq:liku} attains its global maximum. This means that Equation~\eqref{eq:tian} should be violated for at least a value of $\bm{W}^c$ in a c-components. In this case, putting Equation~\eqref{eq:tian} in Equation~\eqref{eq:likdec} produces a value smaller than the global maximum $\lambda^*$, as the maximum is achieved if and only if the values in Equation~\eqref{eq:freqs} are used. The same deduction can be applied to any $\theta_{\bm{U}}$ when $\mathcal{K}=\emptyset$. \qed\\

\paragraph{\bf Proof of Corollary \ref{cor:k}}
The result trivially follows from Theorem~\ref{th:unimod}. \qed\\

\paragraph{\bf Proof of Theorem \ref{th:markov}}
To prove the theorem, let us first take an FSCM $M' \in \mathcal{M}_{M,\tilde{P}}$ and prove that this is also a BN of the CN based on the SEs of M and the CSs returned by Algorithm~\ref{alg:simple}. By the definition of $\mathcal{M}_{M,\tilde{P}}$, $M'$ satisfies Equation~\eqref{eq:ident}, which is equivalent to Equation~\eqref{eq:tian} thanks to Lemma~\ref{lem:tian}. As in a Markovian model $U$ is the unique element of $\bm{U}^c$ and $\mathrm{Pa}(V)=(U,\bm{W}_V)$, this means:
\begin{equation}\label{eq:markovcond2}
\sum_{u\in\Omega_U} P(u) \cdot P(v|u,\bm{w}_V)=P(v|\bm{w}_V)\,,
\end{equation}
and hence:
\begin{equation}\label{eq:markovcond3}
\sum_{u\in\Omega_U} P(u) \cdot \llbracket f_V(u,\bm{w}_V)=v \rrbracket =P(v|\bm{w}_V)\,.
\end{equation}
Equation \eqref{eq:markovcond3} is the constraint in line 4 of Algorithm \ref{alg:simple}. This proves $P(U)\in K(U)$ for each $U\in\bm{U}$, and this means that $\mathcal{M}_{M,\tilde{P}}$ is included in the (set of BNs associated with the) CN.

To prove the inverse inclusion, let us take a PMF $P(U)\in K(U)$ for each $U\in\bm{U}$. This induces a (Markovian) FSCM $M'$ based on M that should satisfy Equation~\eqref{eq:markovcond3} because of the definition of $K(U)$ as in line 4 of Algorithm~\ref{alg:simple}. But this means that also Equation~\eqref{eq:markovcond2} and hence Equation~\eqref{eq:tian} is satisfied. Lemma~\ref{lem:tian} eventually implies $M' \in \mathcal{M}_{M,\tilde{P}}$ and hence the thesis follows. \qed\\

\paragraph{\bf Proof of Theorem \ref{th:hard}}
The proof follows from the analogous result for CNs derived by \citet{cozman2002}. The polytree-shaped CN used in that proof (Figure~\ref{fig:pscm_fabio}) has degenerate CCPTs for the non-root nodes. We can intend those CCPTs as the SEs of a PSCM M and regard the variables associated with the non-root nodes of the CN as the endogenous variables of M. These endogenous nodes form a chain, and each node has a single exogenous parent apart from the first one in the chain, which has also a second exogenous parent. Even if such a model is non-Markovian, we can add an auxiliary exogenous parent to one of the two parents of the first node of the chain and obtain an equivalent Markovian model such that  Algorithm~\ref{alg:simple} returns the CSs on the root node of the polytree-shaped CN.

\begin{figure}[htp!]
\centering
\begin{tikzpicture}[scale=1.]
\node[dot2,label=above:{$U_0$}] (u0)  at (-1,0) {};
\node[dot2,label=above:{$U_1$}] (u1)  at (0,1) {};
\node[dot2,label=above:{$U_2$}] (u2)  at (1,1) {};
\node[dot2,label=above:{$U_n$}] (u4)  at (3,1) {};
\node[dot,label=below:{$V_1$}] (x1)  at (0,0) {};
\node[dot,label=below:{$V_2$}] (x2)  at (1,0) {};
\node[] (x3)  at (2,0) {$\ldots$};
\node[dot,label=below:{$V_n$}] (x4)  at (3,0) {};
\draw[a] (u0) -- (x1);
\draw[a] (u1) -- (x1);
\draw[a] (u2) -- (x2);
\draw[a] (u4) -- (x4);
\draw[a] (x1) -- (x2);
\draw[a2] (x2) -- (x3);
\draw[a2] (x3) -- (x4);
\end{tikzpicture}
\caption{The polytree used in the complexity proof of \citet{cozman2002}.}
\label{fig:pscm_fabio}
\end{figure}

As a query, the authors consider the upper bound of a marginal query in the last node of the chain. The task amounts to the identification of the upper bound of the causal effect on the last node of the chain given an intervention in an additional endogenous parent of the first node of the chain. An algorithm to bound interventional queries in PSCMs would therefore solve inference in polytree-shaped CN. This contradicts the result of the authors. \qed\\

\paragraph{\bf Proof of Theorem \ref{th:quasi}}
For a quasi-Markovian model, Equation~\eqref{eq:tian} rewrites as:
\begin{equation}\label{eq:tian2}
\sum_{u \in \Omega_U} \left[ P(u) \cdot \prod_{V\in\bm{V}^c} P(v|u,\mathrm{pa}_V') \right] = \prod_{V\in\bm{V}^c} P(v|\bm{w}_V)\,,
\end{equation}
and hence:
\begin{equation}\label{eq:tian3}
\sum_{u \in \Omega_U} \left[ P(u) \cdot \prod_{V\in\bm{V}^c} 
 \llbracket f_V(u,\mathrm{pa}_V')=v \rrbracket \right] = \prod_{V\in\bm{V}^c} P(v|\bm{w}_V)\,,
\end{equation}
which coincides with the linear constraint for $P(U)$ in line 6 of Algorithm~\ref{alg:quasi}. Such a remark allows to prove the thesis by a scheme perfectly analogous to that considered for the proof of Theorem~\ref{th:markov}.\qed\\

An alternative proof of Theorem~\ref{th:quasi} based on an arc reversal strategy can be found in \citet{zaffalon2020}.

\paragraph{\bf Proof of Theorem \ref{th:conservative}}
Because of Corollary~\ref{cor:k}, the thesis is equivalent to the fact that the constraints in Equation~\eqref{eq:ident} can be satisfied for at least a specification of $P(\bm{U})$.  Lemma~\ref{lem:tian} allows to replace 
Equation~\eqref{eq:ident} with Equation~\eqref{eq:tian}, to be considered for each $c\in\mathcal{C}$.
As M is Markovian, as already discussed for the proof of Theorem~\ref{th:markov}, we can rewrite Equation~\eqref{eq:tian} as:
\begin{equation}\label{eq:linsys}
\sum_{u\in\Omega_U} P(u) \cdot \llbracket f_V(u,\bm{w}_V)=v \rrbracket =P(v|\bm{w}_V)\,.
\end{equation}
to be satisfied for each $v\in\Omega_V$ and $\bm{w}_V\in\Omega_{\bm{W}_V}$. The representation result provided by \citet[Theorem~1]{druzdzel1993causality} can be used to prove that the linear constraints in Equation~\eqref{eq:linsys} can be always satisfied if the SEs are canonical. In their proof the authors consider a setup analogous to the current one but, instead of $U$, a continuous $U'\in[0,1]$ with a uniform density $P(U')$ is considered. Say that $\Omega_V := \{v_1,\ldots,v_q\}$. For each $\bm{w}_V\in\Omega_{\bm{W}_V}$, define the vector $H_{\bm{w}_V}:=\{h_{\bm{w}_V}^{(i)}\}_{i=0}^{q}$ such that $h_{\bm{w}_V}^{(0)}:=0$ and $h_{\bm{w}_V}^{(i)}:= \sum_{j=1}^i P(v_j|\bm{w}_V)$ for each $i=1,\ldots,q$.
Note that $h_{\bm{w}_V}^{(i)}\leq h_{\bm{w}_V}^{(i+1)}$ for each $i=1,\ldots,q-1$ and $h_{\bm{w}_V}^{(q)}=1$. The authors show that Equation~\eqref{eq:linsys} is satisfied if the SE $V=f_V'(U',\bm{W}_V)$ is such that:
\begin{equation}\label{eq:fprime}
f_V'(u',\bm{w}_V) := \left\{ v_i \in \Omega_V : u' \in \left[
h_{\bm{w}_V}^{(i)},h_{\bm{w}_V}^{(i+1)} \right] \right\}\,,
\end{equation}
for each $u' \in [0,1]$ and $\bm{w}_V\in\Omega_{\bm{W}_V}$. 
A partition of $[0,1]$ is obtained by
removing the left endpoints from the intervals in Equation~\eqref{eq:fprime}  apart from the first one. Thus, for a given $\bm{w}_V\in\Omega_{\bm{W}_V}$, Equation~\eqref{eq:fprime} can be regarded as a discrete SE, mapping to $V$ the values of a discretisation of $U'$ based on the partition of $[0,1]$ induced by $H_{\bm{w}_V}$. 

A \emph{least common partition} is obtained from a set of partitions by putting together and sorting the endpoints of the intervals of all the partitions. Take the discretisation of $U'$ induced by such a least common partition when considering 
all the partitions of $[0,1]$ induced by each $\bm{w} \in \Omega_{\bm{W}_V}$.
In practice, the SE $V=f_V'(\bm{W}_V,U')$ can be equivalently described by $V=\hat{f}_V(\bm{W}_V,\hat{U})$, where $\hat{U}$ is a discrete variable whose states are in correspondence with the above considered set of discretisation intervals for $U$. The uniform density $P(U')$ is consequently mapped to a PMF $P(\hat{U})$ such that $P(\hat{U}=\hat{u})$ is equal to the integral of $P(U')$ on the interval associated with $\hat{u}$ and hence it is equal to its width. Finally, observe that, for each $\hat{u}\in\Omega_{\hat{U}}$, $\hat{f}_V$ defines a deterministic relation between $\bm{W}_V$ and $\bm{V}$ and this should correspond to a state of $U$ in M, as $U$ is enumerating all these possible relations because of the canonical specification. This defines a map $\mu:\Omega_{\hat{U}}\rightarrow\Omega_U$ with $\mu(\hat{u}):=\{ u \in \Omega_U: f_V(u,\bm{W}_V) = \hat{f}(\hat{u},\bm{W}_V)\}$. For $P(U)$ we have:
\begin{equation}
P(u) = \sum_{\hat{u} \in \Omega_{\hat{U}}: \gamma(\hat{u})=u} P(\hat{u})\,,
\end{equation}
and $P(u)=0$ for the states of $u$ that are not in the domain of $\mu$. This is the exogenous quantification that proves the thesis. \qed\\

\paragraph{\bf Proof of Theorem \ref{cor:discarding}}
Note that $M$ embeds $M'$ if and only if there is at least a $P(\bm{U})\in\mathcal{K}$ so that $P(U \in \Omega_U')=1$, for all exogenous variables $U$.

Assume, \emph{ad absurdum}, that $M$ embeds an incompatible $M'$. For each $U\in\bm{U}$, let $\mathcal{K}'(U)$, $U\in\bm{U}$, denote the sets of compatible specifications of $M'$ relative to variable $U$. By Corollary~\ref{cor:k}, the incompatibility of $M'$ implies that $\mathcal{K}'(U)$ should be empty for at least a $U\in\bm{U}$. By the definition of embedding, for each $U\in\bm{U}$, we should have at least a $P(U)\in \mathcal{K}(U)$ such that $P(U \in \Omega_U')=1$.

Obtaining $M'$ from $M$ can be regarded as the result of conditioning the exogenous PMFs on the events $U\in\Omega_U'$, for each $U\in\bm{U}$. As $P(U \in \Omega_U')=1$, such a conditioning is well defined for $P$. Let $P'(U)$ denote the resulting PMF for $M'$, which simply corresponds to the restriction of $P$ to $\Omega_U'$.

Consider the compatibility constraints of $M$ involving PMF $P(U)$ as in Equation~\eqref{eq:ident}. As $P(U) \in \mathcal{K}(U)$, these constraints should be satisfied by $P(U)$. With Markovian models, the constraints are linear. Since by $P(U \in \Omega_U')=1$, the same constraints hold for $P'(U)$. But this means $P'(U) \in\mathcal{K}'(U)$ and hence $\mathcal{K}'(U) \neq \emptyset$ (for each $U\in\bm{U}$), which is a contradiction. \qed\\




\paragraph{\bf Proof of Theorem \ref{th:newbounds}}
Consider the l.h.s. of Equation~\eqref{eq:confidenceBeta}. The corresponding joint density is:
\begin{equation}
P\left( \Delta_a \leq \dfrac{\delta_a}{2}, \Delta_b \leq \dfrac{\delta_b}{2}, \rho \right) = \gamma \int_{0}^{\frac{\delta_b}{2}}\int_{0}^{\frac{\delta_a}{2}} P(\rho | \Delta_a = x, \Delta_b = y) \mathrm{d}x \mathrm{d}y\,,
\end{equation}
where a uniform prior is considered for $\Delta_a := (a-a^*)$ and $\Delta_b := (b^*-b)$. 

As $\pi_i \sim \operatorname{Beta}(\alpha, \beta, a^*, b^*)$:
\begin{equation}
P(\pi_i \in [a,b]|\Delta_a=x,\Delta_b=y)=P(\pi_i \in [a,b]|a^*=a-x,b^*=b+y)\,,
\end{equation}
and hence:
\begin{equation}
P(\pi_i \in [a,b]|\Delta_a=x,\Delta_b=y)=
\int_{a}^{b} \dfrac{(\rho-a+x)^{\alpha-1}(b+y-\rho)^{\beta-1}}{(b-a+y+x)^{\alpha+\beta-1}B(\alpha,\beta)} \mathrm{d}\rho\,.
\label{eq:intRho}
\end{equation}
We solve~\eqref{eq:intRho} analytically and obtain that $P(\rho|\Delta_a=x,\Delta_b=y)$ equals to:
\begin{equation}\footnotesize
\left(\dfrac{(b-a+x)^\alpha \twoFone(\alpha,1-\beta,\alpha+1,\frac{b-a+x}{b-a+x+y})- x^\alpha \twoFone(\alpha,1-\beta,\alpha+1,\frac{x}{b-a+x+y})}{\alpha (b-a+x+y)^{\alpha} B(\alpha,\beta)}\right)^k\,.
\end{equation}
The joint $P(\rho, \Delta_a\leq \frac{\delta_a}{2},\Delta_b\leq \frac{\delta_b}{2})$ can thus be obtained by the following integral:
\begin{equation}
\footnotesize
\gamma \!\!\int_{0}^{\delta_b/2}\!\!\!\int_{0}^{\delta_a/2}\!\!\left(\dfrac{(b-a+x)^\alpha \twoFone(\alpha,1-\beta,\alpha+1,\frac{b-a+x}{b-a+x+y})- x^\alpha \twoFone(\alpha,1-\beta,\alpha+1,\frac{x}{b-a+x+y})}{\alpha (b-a+x+y)^{\alpha} B(\alpha,\beta)}\right)^k
\!\!\!
\mathrm{d}x \mathrm{d}y
\,.
\label{eq:intAppTh8}
\end{equation}
Note that the condition on the l.h.s. of Equation~\eqref{eq:confidenceBeta} is $\Delta_a=a-a^* \leq \varepsilon L$, but the condition in Equation~\eqref{eq:confidenceBeta} is only meaningful if $\Delta_a \geq 0$, i.e. if $a \geq \varepsilon L$. This means that $\frac{\delta_a}{2} = \varepsilon L$ if $a \geq \varepsilon L$ and $\frac{\delta_a}{2} = a$ otherwise. By the same reasoning on $\Delta_b$, if we rearrange the conditions we obtain the values for $\delta_a, \delta_b$ outlined in Equations~\eqref{eq:delta_a} and \eqref{eq:delta_b}. 

The marginal distribution for $\rho$ can also be obtained by solving the following integral:
\begin{align*}
P(\rho)= \gamma \int_{0}^{a+(1-b)}\int_{0}^{a+(1-b)-y}P(\rho | \Delta_a=x,\Delta_b=y) \mathrm{d}x \mathrm{d}y\,.
\end{align*}
The l.h.s. of Equation~\eqref{eq:confidenceBeta} is just the ratio between $P(\rho, \Delta_a\leq \frac{\delta_a}{2},\Delta_b\leq \frac{\delta_b}{2})$ and $P(\rho)$.\qed\\

\paragraph{\bf Proof of Corollary \ref{cor:boundaequalb}}
Let us discuss separately the three cases.

\emph{Case $a=0$.} If $a=0$ then $a^*=0$ and $\Delta_a=0$, then $P(\pi_i \in [a,b] | a^*=0, b^*=b+y)$ becomes
\begin{equation}
    \footnotesize
\int_{a}^{b} \dfrac{(\rho)^{\alpha-1}(b+y-\rho)^{\beta-1}}{(b+y)^{\alpha+\beta-1}B(\alpha,\beta)} \mathrm{d}\rho\,.
\end{equation}
This integral can be solved analytically and, by exploiting the independence of $\pi_i$, we obtain the expression in Equation~\eqref{eq:newbounds3}. By following the same steps as in the proof of Theorem~\ref{th:newbounds} we obtain the result.

\emph{Case $b=1$.} If $b=1$, then $b^*=1$ and $\Delta_b=0$, then $P(\pi_i \in [a,b] | a^*=a-x,b^*=1)$ becomes 
\begin{equation}\footnotesize
\int_{a}^{b} \dfrac{(\rho-a+x)^{\alpha-1}(1-\rho)^{\beta-1}}{(1-a+x)^{\alpha+\beta-1}B(\alpha,\beta)} \mathrm{d}\rho\,.
\end{equation}
This integral can be solved analytically and, by exploiting the independence of $\pi_i$, we obtain the expression in Equation~\eqref{eq:newbounds2}. By following the same steps as in the proof of Theorem~\ref{th:newbounds} we obtain the result.

\emph{Case $a=b$.} If $a=b$ then $a^*=b^*$ and $L\rightarrow 0$. Moreover we assume here that $\delta \rightarrow 0$ and $\epsilon \rightarrow 1$. We assume that the distribution of the outputs of the EMCC iterations is uniformly distributed in  $[a^*,b^*]$. $P(a=b |\rho)$ is obtained from Equation~\eqref{eq:bounds_uniform} in the limit $L\to 0$ and $\epsilon \to 1$ and the thesis follows trivially.\qed\\

\paragraph{\bf Proof of Corollary \ref{cor:newbounds}}
For $\alpha=\beta=1$, $\twoFone(\alpha,1-\beta,\alpha+1,\mu)=1$, $B(\alpha,\beta)=1$ and hence:
\begin{equation}
P(\rho |\Delta_a=x,\Delta_b=y)=
\left(\dfrac{b-a+x- x}{b-a+x+y}\right)^n = \left(\dfrac{b-a}{b-a+x+y}\right)^k\,. 
\end{equation}
Equation~\eqref{eq:bounds_uniform} is finally obtained by computing the integrals as in \citet{zaffalon2021}.\qed\\

{\color{myblueR2}{\section{Characterising the Likelihood in the Joint Case}\label{app:converg}
Equation~\eqref{eq:liku} contains the product of the chances of the exogenous variables in $\bm{U}$; this is consequence of their being mutually independent (that is, being root nodes in an SCM). In this appendix, we study the log-likelihood function in the simpler setting in which we work in the space of the joint chances $\theta_{\bm{U}}$. This is equivalent to considering an SCM with a single exogenous node; we call the resulting log-likelihood $l'$. The latter contains the linear term $\theta_{\bm{U}}$ where $l$ had instead the product of the chances. This makes $l'$ easier to characterise than $l$, in the way that follows (we assume we are under M-compatibility to make sure that a global maximum exists).

\begin{lemma}\label{lem:concave}
The function $l'$ is concave. 
\begin{proof}
Consider a dataset made of a single observation. In this case, $l'$ is the logarithm of a linear function of $\theta_{\bm{U}}$, whence it is concave. For more general datasets, $l'$ is then a sum of concave functions, which is also concave.
\end{proof}
\end{lemma}

\begin{theorem}\label{th:saddle}
The function $l'$ cannot have saddle points or local maxima.
\begin{proof}
This is a direct consequence of Lemma~\ref{lem:concave}.
\end{proof}
\end{theorem}

\begin{theorem}\label{th:convex}
Denote by $l^*$ the global maximum value of $l'$. The global maximum points of $l'$ form a convex region, i.e., for any two points $\theta_{\bm{U}}',\theta_{\bm{U}}''$ such that $l'(\theta_{\bm{U}}')=l'(\theta_{\bm{U}}'')=l^*$, then $l'(\theta_{\bm{U}})=l^*$ for each $\theta_{\bm{U}}:=\alpha \theta_{\bm{U}}' + (1-\alpha) \theta_{\bm{U}}''$ with $0 \leq \alpha \leq 1$.
\begin{proof}
Lemma~\ref{lem:concave} implies $l'(\theta_{\bm{U}}) \geq \alpha l'(\theta_{\bm{U}}')+(1-\alpha)l'(\theta_{\bm{U}}'')$ and hence $l'(\theta_{\bm{U}}) \geq l^*$. But, as the maximum of $l'(\theta_{\bm{U}})$ is $l^*$, we have $l'(\theta_{\bm{U}})=l^*$, i.e., the thesis.
\end{proof}
\end{theorem}

The characterisations above are not immediate to extend to the case of $l$, due to the non-linearity introduced by the product of the chances. However we conjecture that in such a case there should not be local maxima either, since we have not encountered that case in our experiments. We leave this proof for future work.}}

\section{Questions and Answers}\label{app:faq}
In this section, we discuss various aspects of the paper in an attempt to clarify them via a more direct approach. We are taking inspiration from our rebuttal to the reviewers' reports to lay down a section that should hopefully address the main doubts a reader might be concerned with.

\begin{quote}{\bf Q1}.
Are the bounds you compute credible intervals? That is, are you after the `true' distribution obtained in the limit of infinite data? What do you mean by saying that bounds are `exact'? \end{quote}

This is perhaps one of the biggest potential sources of confusion, therefore we will discuss it at some length.

Let us consider for a moment a Bayesian network. When we estimate its parameters from data, however we do that, we are after some kind of, frequentist or Bayesian or some other type of, expectation (in particular, the expectation of indicator functions $I$). We plug these expectations into the network as parameters and then we use it to do inference, like belief updating or \emph{maximum a posteriori} explanation. We typically do not use credible intervals or the like to estimate the net's parameters, we just use expectations.

Now imagine that the Bayesian net embeds one latent root node $U$. We still want to estimate the net's parameters from data; but we have data only about the manifest variables. In this case we typically use the EM or some other algorithms to estimate the parameters of the unobserved node, that is, its unconditional probability mass function $P(U)$. What we get in the end is again an  expectation $E[I_{U=u}]=P(U=u)$, for all values $u$ that $U$ can take on. And in particular, if the problem is identifiable, the estimated $P(U)$ corresponds to the global maximum $\theta^*_{\mathbf{U}}$ of the EM. The point is that also in this case, we are after expectations.

Now consider the case of unidentifiable problems. In this case there are multiple unconditional mass functions $P(U)$ for the latent node that lead to the distribution we estimate from data (the empirical distribution) once we marginalise $U$ out. Each one of them corresponds to a global maximum of the EM (a result in this sense was already given by \citealt{redner1981}). We take the set of all these unconditional mass functions and call it $\mathcal{K}:=\{P(U)\}$. We have no information as to whether one element of $\mathcal{K}$ is more probable than another. We just have no second order probability on top of $\mathcal{K}$---which is the essence of non-identifiability.

The best we can do, at this point, is to consider the (possibly infinite) set of Bayesian networks that we obtain by considering all the mass functions for $U$ that are in $\mathcal{K}$; we run our inference (say, updating) on each of them and we summarise the results by the lower and the upper values of the updating over such a set of Bayesian nets. These are unidentifiabilty bounds; all we know is where the updated probability lies (we have no probability distribution on top of such a probability, even less so a uniform one). Let us stress that those are indeed bounds on an expectation (of an indicator function). They are not determining a credible interval. More specifically, those bounds have nothing to do with inferring population values from sample outcomes. The procedure is just the same that we follow for Bayesian nets without latent nodes; we estimate expectations and deliver expectations. The difference is `only' that now we have interval-valued expectations, because of unidentifiability (which means that our model is incomplete). What we are saying so far is that non-identifiability has not necessarily to be subject to question of statistical inference; it is rather like identifiable problems: we use our models via expectations.

And here is the crucial point of potential confusion: whenever in the paper we refer to the `exact' bounds, we refer to the interval-valued expectation that we would obtain if we could run all the models in $\mathcal{K}$ one after the other and then summarise the (observational, interventional, counterfactual) results by the lower and upper bounds obtained across all those models. Usually we rather try to compute directly exact bounds via an optimisation problem---which is our credal-net based approach in this paper. But that is not possible in general and we need approximations. In this case we resort to our EMCC, which will deliver ranges, i.e., inner approximations  of the bounds.

Let us stress once more that we never aim in this paper to compute the counterfactual bounds that we would get if we had an infinite sample to estimate parameters. In the same way as in Bayesian nets one does not refer to the outcome of an updating as the actual value of the probability that one would obtain if the parameters of the Bayesian nets were estimated form an infinite sample. We just know this is impossible and we stick to the best we can: using expectations. For us `exact' refers to exactness with respect to optimising over $\mathcal{K}$.

An additional note is that also our credible intervals should be understood in the same light. That is, we yield an approximating interval-valued expectation (probability) $[a,b]$, which we know that by construction is included in the exact interval-valued expectation $[a^*,b^*]$: where the latter is made by the exact yet unknown bounds that result from an optimisation problem over $\mathcal{K}$. Our credible interval tells us then, in probability, how much smaller is $[a,b]$ compared to $[a^*,b^*]$. Note that the former gets closer and closer to the latter the more EM runs we do, with the same data fixed (not with increasing data).

\begin{quote}{\bf Q2}.
M-compatibility seems to be referring to the empirical distribution. Shouldn't it refer to the `true' one? Moreover, the empirical distribution is estimated via maximum likelihood and obtained via the factorisation in~\eqref{eq:empirical}. Is your notion of compatibility restricted only to these specific choices?
\end{quote}

Our definition of compatibility indeed applies to the empirical distribution; this paper is not focused on statistical inference. It is true that there are different ways to define the empirical distribution (we use MLE, but one could use Bayes; we use (4)---because that is induced by the very marginal DAG that generates the data---, but others might want to use other factorisations); and yet we are not concerned about the way by which one defines the empirical distribution. What matters to us is that once it is defined, it complies with the logical constraints imposed by the SCM, otherwise we would be using a model that embeds a logical contradiction and that would not allow us to make sensible calculations. 

Let us also note that defining compatibility with the `true' distribution (obtained in the limit of infinite data) can be questionable, because the compatibility problem vanishes in such a case: in fact, one could use a canonical specification (that is always compatible with the data---as we show for Markovian SCMs), thus eventually (i.e., in the limit) automatically retaining only the distributions that are compatible with the given causal graph.

\begin{quote}
{\bf Q3}. What is Theorem~\ref{th:unimod} actually proving?
\end{quote}

We start by having assessed in some way the empirical mass function $P({\bf V})$. If our problem were identifiable, the global maximum of the EM would correspond to the $P({\mathbf U})$ that leads to $P({\bf V})$ when the $U$ variables are marginalised out of the given SCM. In reality the problem is not identifiable and hence there are (possibly infinitely) many $P({\mathbf U})$ that lead to $P({\bf V})$ when we marginalise the $U$ variables out. The theorem's claim is just that those many $P({\mathbf U})$, which we gather in set $\mathcal{K}$ (representing them as collections of marginals over the exogenous nodes), are in a one-to-one correspondence with the global maxima of the likelihood.

Stated differently, the problem of the logical compatibility of the structural equations with a given (empirical) probability distribution, can be regarded as one of feasibility: that is, we could write it down as the problem determining whether the feasible region of an optimisation problem is empty or not. The theorem allows us to sample such a region by multiple runs of the EM (instead of solving the NP-hard optimisation problem).

\begin{quote}
{\bf Q4}. The paper advocates testing for M-compatibility, which is a valid point; Examples~\ref{ex:pearl}--\ref{ex:pearl4} are interesting and useful. But do they justify the claims the paper makes, such as:
\begin{itemize}
\item if the true underlying model is not compatible with the available data, the results obtained by using the canonical model will be unwarranted as an approximation to the actual one.
\item We cannot have guaranteed bounds without knowing the structural equations of the underlying SCM.
\end{itemize}
Does estimated $P$ being compatible with $M$ or not really make a fundamental difference (like a phase transition) in the validity of the inferred bounds? Or the quality of the inferred bounds simply depends on the distance between $P$  and the `true' distribution of the endogenous variables? Is it just a problem of statistical inference then?
\end{quote}
Let us stay for a moment in the case of finite sample sizes. Our point is that many results published for that case assume that a causal graph (that is, one without structural equation) together with data about the exogenous nodes can yield us valid bounds. Our examples show that this is not the case; it is particularly disturbing that the empirical distribution can be incompatible with the very SCM that has generated it.

Part of the questions asked seems to imply the following: can this be avoided with your notion of compatibility? In a sense, yes: let us imagine that we can test all the PSCMs with the given causal graph, and that each of them is compatible with the empirical distribution. This test makes sure that our expectation bounds (which are of the same type of those by~\citealt{exact2021causes}, for instance) make sense: because  they have passed a basic test of rationality. In fact, even if only one of those PSCMs were incompatible, we would be leaving open the possibility that such a PSCM is the actual one generating the data; and our bounds would be at risk of being wrong, because they would not cover such a case (as in our examples).

One point that we make is that the canonical specification is not a way out to such a problem. In principle, it seems to allow us to use all the possible SEs at once (which is also the reason why it is always compatible with the empirical distribution), thus bypassing the problem. But in reality the incompatible SEs/PSCMs (implicitly considered in the canonical specification) are just put under the carpet by internally never assigning probability zero to the structural equations that do not belong to them, which is equivalent to actually neglecting those sub-models; and the problem would still be there.

Yet a sufficient condition to address the problem exists: we should use the canonical specification and check whether all sub-models are embedded. In the Markovian case this is readily possible to do since we can exactly represent the set $\mathcal{K}$ via linear constraints as Theorem~\ref{th:markov} proves (with an increase in complexity, the same holds for quasi-Markovian models). In case they were, our expectation bounds would not be at risk. In the opposite case, we should give the disclaimer that the bounds are not guaranteed to be correct; or we should just refrain from yielding them.

Now let us consider inferential arguments, that is, how this relates to the similarity of the `true' distribution to the empirical one. Let us consider again the canonical specification. Since this implicitly considers all the possible SEs, in the limit of infinite data, all and only the compatible models will be embedded. And since the true SCM that has generated the data must be compatible with the `true' distribution, by definition, it will be included among them, and our inferences will be correct. 

The problem here is that all this happens in the limit, and we do not know yet how this should be related to questions of finite samples: how large will a sample have to be to declare that the true SCM is for sure embedded?

We seem to have to connect the distance between the distributions to the transition into compatibility of the true SCM. But we do not know whether this is a continuous map, and there could indeed be a kind of phase-transition (in particular because SEs are deterministic, sharp, equations). As long as this map is not studied in detail, we do not seem to be in the conditions to give probabilities of compatibility along these lines.

\begin{quote}
{\bf Q5}. One stated advantage of the proposed EMCC approach over the existing work is that it provides ranges that approximate the exact ones from inside (i.e., a so-called inner approximation). Where does this claim comes from? 

It looks as if $[a^*,b^*]$ here are not the `true' exact bounds due to partial identifiability but those one obtains by assuming $P$ in~\eqref{eq:freqs} is the `true' distribution. Thus $[a^*, b^*]$ approach the `true' bounds only given infinite samples; otherwise, their relation with the `true' bounds is unclear as well as it is unclear how EMCC can escape the finite sample issue.
\end{quote}

The question in the first paragraph relates to some of our previous replies: `exact' bounds does not mean in this paper the bounds one gets with infinite data; rather it is the exact solution of an optimisation problem over $\mathcal{K}$. In the second paragraph the question is similar:  we do not aim at escaping the finite sample issue by the EMCC.

\begin{quote}
{\bf Q6}. It looks as if EMCC vs. \citet{zhang2021} is basically frequentist vs. Bayesian. 
\end{quote}

Actually we would not say that this is the main issue at stake. Let us try to clarify the main differences between our EMCC and the alternative sampling method proposed by \citet{zhang2021} to approximate partially identifiable queries. 

The latter is indeed based on Bayesian ideas, in the sense that a prior is put over the chances of the exogenous variables that is later updated to a posterior via the dataset 
$\mathcal{D}$ of endogenous observations. With the posterior they can get an expected value over the chances that in turn leads to an expected value over the query of interest. That is to say that the partially identifiable counterfactual query is reduced to a point estimate when it comes to expectation. This should be constrasted with the EMCC, which delivers approximate lower and upper expectations (bounds) for the query of interest. 

However \citet{zhang2021} are after the expectation bounds as we do. To this end, they use a different strategy than ours and that is based on credible intervals. Their Theorem 3.2 shows that the 100\% credible interval contains the expectation bounds (i.e., it is an \emph{outer} approximation) and coincides with it almost surely in the limit. When dealing with finite samples, they provide further results giving probabilistic guarantees that the computed interval is in fact a 100\% credible interval. This result is based on a tolerance rate $\varepsilon$ that is somewhat similar in its aim to the $\varepsilon$ we use in Section~\ref{sec:credible} for our own credible intervals. In fact the two, quite different, types of credible intervals, theirs and ours, can be regarded to be after the same goal: that is, to state with which probability the delivered interval contains the exact expectation interval.

It would be interesting in the future to compare the relative power of the two types of credible intervals. For the time being, let us just note that the EMCC provides an inner approximation to the actual expectation bounds, besides a probabilistic guarantee on the fact that the credible interval is an outer approximation. Furthermore, \citet{zhang2021} appears to be oblivious to the question of M-compatibility, which, as we saw, has the power to invalidate what might look like safe conclusions.

\section*{Acknowledgements}
\indent The reviewers of this paper have been very careful in checking it, thus helping us to identify some of its unclear and imprecise parts; we gratefully acknowledge their dedication and stubbornness, which the paper has definitely benefited from. We are also grateful to Heidi Kern from the Triangolo association for her support with the palliative care problem discussed in Section~\ref{sec:triangolo}. This research was partially funded by MCIN/AEI/10.13039/501100011033 with FEDER funds for the projects PID2019-106758GB-C32 and PID2022-139293NB-C31. Finally, we would like to thank the ``Mar\'{i}a Zambrano'' grant (RR\_C\_2021\_01) from the Spanish Ministry of Universities and funded with NextGenerationEU funds.

\end{document}